\documentclass[runningheads]{llncs}
\usepackage{graphicx}
\usepackage{amsmath,amssymb} 
\usepackage{color}

\usepackage{graphicx}
\usepackage{comment}
\usepackage{amsmath,amssymb} 
\usepackage{color}
\usepackage{caption}
\usepackage{subcaption}
\usepackage[width=122mm,left=12mm,paperwidth=146mm,height=193mm,top=12mm,paperheight=217mm]{geometry}

\usepackage{amssymb}
\usepackage{pifont}
\newcommand{\cmark}{\ding{51}}%
\newcommand{\xmark}{\ding{55}}%
\usepackage{floatrow}
\usepackage{floatrow}
\usepackage{setspace}
\usepackage{ulem}
\usepackage{xstring}
\usepackage{footnote}
\usepackage{color}

\usepackage[draft]{minted}
\makeatletter
\def\PYGdefault@reset{\let\PYGdefault@it=\relax \let\PYGdefault@bf=\relax%
    \let\PYGdefault@ul=\relax \let\PYGdefault@tc=\relax%
    \let\PYGdefault@bc=\relax \let\PYGdefault@ff=\relax}
\def\PYGdefault@tok#1{\csname PYGdefault@tok@#1\endcsname}
\def\PYGdefault@toks#1+{\ifx\relax#1\empty\else%
    \PYGdefault@tok{#1}\expandafter\PYGdefault@toks\fi}
\def\PYGdefault@do#1{\PYGdefault@bc{\PYGdefault@tc{\PYGdefault@ul{%
    \PYGdefault@it{\PYGdefault@bf{\PYGdefault@ff{#1}}}}}}}
\def\PYGdefault#1#2{\PYGdefault@reset\PYGdefault@toks#1+\relax+\PYGdefault@do{#2}}

\expandafter\def\csname PYGdefault@tok@gd\endcsname{\def\PYGdefault@tc##1{\textcolor[rgb]{0.63,0.00,0.00}{##1}}}
\expandafter\def\csname PYGdefault@tok@gu\endcsname{\let\PYGdefault@bf=\textbf\def\PYGdefault@tc##1{\textcolor[rgb]{0.50,0.00,0.50}{##1}}}
\expandafter\def\csname PYGdefault@tok@gt\endcsname{\def\PYGdefault@tc##1{\textcolor[rgb]{0.00,0.27,0.87}{##1}}}
\expandafter\def\csname PYGdefault@tok@gs\endcsname{\let\PYGdefault@bf=\textbf}
\expandafter\def\csname PYGdefault@tok@gr\endcsname{\def\PYGdefault@tc##1{\textcolor[rgb]{1.00,0.00,0.00}{##1}}}
\expandafter\def\csname PYGdefault@tok@cm\endcsname{\let\PYGdefault@it=\textit\def\PYGdefault@tc##1{\textcolor[rgb]{0.25,0.50,0.50}{##1}}}
\expandafter\def\csname PYGdefault@tok@vg\endcsname{\def\PYGdefault@tc##1{\textcolor[rgb]{0.10,0.09,0.49}{##1}}}
\expandafter\def\csname PYGdefault@tok@m\endcsname{\def\PYGdefault@tc##1{\textcolor[rgb]{0.40,0.40,0.40}{##1}}}
\expandafter\def\csname PYGdefault@tok@mh\endcsname{\def\PYGdefault@tc##1{\textcolor[rgb]{0.40,0.40,0.40}{##1}}}
\expandafter\def\csname PYGdefault@tok@go\endcsname{\def\PYGdefault@tc##1{\textcolor[rgb]{0.53,0.53,0.53}{##1}}}
\expandafter\def\csname PYGdefault@tok@ge\endcsname{\let\PYGdefault@it=\textit}
\expandafter\def\csname PYGdefault@tok@vc\endcsname{\def\PYGdefault@tc##1{\textcolor[rgb]{0.10,0.09,0.49}{##1}}}
\expandafter\def\csname PYGdefault@tok@il\endcsname{\def\PYGdefault@tc##1{\textcolor[rgb]{0.40,0.40,0.40}{##1}}}
\expandafter\def\csname PYGdefault@tok@cs\endcsname{\let\PYGdefault@it=\textit\def\PYGdefault@tc##1{\textcolor[rgb]{0.25,0.50,0.50}{##1}}}
\expandafter\def\csname PYGdefault@tok@cp\endcsname{\def\PYGdefault@tc##1{\textcolor[rgb]{0.74,0.48,0.00}{##1}}}
\expandafter\def\csname PYGdefault@tok@gi\endcsname{\def\PYGdefault@tc##1{\textcolor[rgb]{0.00,0.63,0.00}{##1}}}
\expandafter\def\csname PYGdefault@tok@gh\endcsname{\let\PYGdefault@bf=\textbf\def\PYGdefault@tc##1{\textcolor[rgb]{0.00,0.00,0.50}{##1}}}
\expandafter\def\csname PYGdefault@tok@ni\endcsname{\let\PYGdefault@bf=\textbf\def\PYGdefault@tc##1{\textcolor[rgb]{0.60,0.60,0.60}{##1}}}
\expandafter\def\csname PYGdefault@tok@nl\endcsname{\def\PYGdefault@tc##1{\textcolor[rgb]{0.63,0.63,0.00}{##1}}}
\expandafter\def\csname PYGdefault@tok@nn\endcsname{\let\PYGdefault@bf=\textbf\def\PYGdefault@tc##1{\textcolor[rgb]{0.00,0.00,1.00}{##1}}}
\expandafter\def\csname PYGdefault@tok@no\endcsname{\def\PYGdefault@tc##1{\textcolor[rgb]{0.53,0.00,0.00}{##1}}}
\expandafter\def\csname PYGdefault@tok@na\endcsname{\def\PYGdefault@tc##1{\textcolor[rgb]{0.49,0.56,0.16}{##1}}}
\expandafter\def\csname PYGdefault@tok@nb\endcsname{\def\PYGdefault@tc##1{\textcolor[rgb]{0.00,0.50,0.00}{##1}}}
\expandafter\def\csname PYGdefault@tok@nc\endcsname{\let\PYGdefault@bf=\textbf\def\PYGdefault@tc##1{\textcolor[rgb]{0.00,0.00,1.00}{##1}}}
\expandafter\def\csname PYGdefault@tok@nd\endcsname{\def\PYGdefault@tc##1{\textcolor[rgb]{0.67,0.13,1.00}{##1}}}
\expandafter\def\csname PYGdefault@tok@ne\endcsname{\let\PYGdefault@bf=\textbf\def\PYGdefault@tc##1{\textcolor[rgb]{0.82,0.25,0.23}{##1}}}
\expandafter\def\csname PYGdefault@tok@nf\endcsname{\def\PYGdefault@tc##1{\textcolor[rgb]{0.00,0.00,1.00}{##1}}}
\expandafter\def\csname PYGdefault@tok@si\endcsname{\let\PYGdefault@bf=\textbf\def\PYGdefault@tc##1{\textcolor[rgb]{0.73,0.40,0.53}{##1}}}
\expandafter\def\csname PYGdefault@tok@s2\endcsname{\def\PYGdefault@tc##1{\textcolor[rgb]{0.73,0.13,0.13}{##1}}}
\expandafter\def\csname PYGdefault@tok@vi\endcsname{\def\PYGdefault@tc##1{\textcolor[rgb]{0.10,0.09,0.49}{##1}}}
\expandafter\def\csname PYGdefault@tok@nt\endcsname{\let\PYGdefault@bf=\textbf\def\PYGdefault@tc##1{\textcolor[rgb]{0.00,0.50,0.00}{##1}}}
\expandafter\def\csname PYGdefault@tok@nv\endcsname{\def\PYGdefault@tc##1{\textcolor[rgb]{0.10,0.09,0.49}{##1}}}
\expandafter\def\csname PYGdefault@tok@s1\endcsname{\def\PYGdefault@tc##1{\textcolor[rgb]{0.73,0.13,0.13}{##1}}}
\expandafter\def\csname PYGdefault@tok@sh\endcsname{\def\PYGdefault@tc##1{\textcolor[rgb]{0.73,0.13,0.13}{##1}}}
\expandafter\def\csname PYGdefault@tok@sc\endcsname{\def\PYGdefault@tc##1{\textcolor[rgb]{0.73,0.13,0.13}{##1}}}
\expandafter\def\csname PYGdefault@tok@sx\endcsname{\def\PYGdefault@tc##1{\textcolor[rgb]{0.00,0.50,0.00}{##1}}}
\expandafter\def\csname PYGdefault@tok@bp\endcsname{\def\PYGdefault@tc##1{\textcolor[rgb]{0.00,0.50,0.00}{##1}}}
\expandafter\def\csname PYGdefault@tok@c1\endcsname{\let\PYGdefault@it=\textit\def\PYGdefault@tc##1{\textcolor[rgb]{0.25,0.50,0.50}{##1}}}
\expandafter\def\csname PYGdefault@tok@kc\endcsname{\let\PYGdefault@bf=\textbf\def\PYGdefault@tc##1{\textcolor[rgb]{0.00,0.50,0.00}{##1}}}
\expandafter\def\csname PYGdefault@tok@c\endcsname{\let\PYGdefault@it=\textit\def\PYGdefault@tc##1{\textcolor[rgb]{0.25,0.50,0.50}{##1}}}
\expandafter\def\csname PYGdefault@tok@mf\endcsname{\def\PYGdefault@tc##1{\textcolor[rgb]{0.40,0.40,0.40}{##1}}}
\expandafter\def\csname PYGdefault@tok@err\endcsname{\def\PYGdefault@bc##1{\setlength{\fboxsep}{0pt}\fcolorbox[rgb]{1.00,0.00,0.00}{1,1,1}{\strut ##1}}}
\expandafter\def\csname PYGdefault@tok@kd\endcsname{\let\PYGdefault@bf=\textbf\def\PYGdefault@tc##1{\textcolor[rgb]{0.00,0.50,0.00}{##1}}}
\expandafter\def\csname PYGdefault@tok@ss\endcsname{\def\PYGdefault@tc##1{\textcolor[rgb]{0.10,0.09,0.49}{##1}}}
\expandafter\def\csname PYGdefault@tok@sr\endcsname{\def\PYGdefault@tc##1{\textcolor[rgb]{0.73,0.40,0.53}{##1}}}
\expandafter\def\csname PYGdefault@tok@mo\endcsname{\def\PYGdefault@tc##1{\textcolor[rgb]{0.40,0.40,0.40}{##1}}}
\expandafter\def\csname PYGdefault@tok@kn\endcsname{\let\PYGdefault@bf=\textbf\def\PYGdefault@tc##1{\textcolor[rgb]{0.00,0.50,0.00}{##1}}}
\expandafter\def\csname PYGdefault@tok@mi\endcsname{\def\PYGdefault@tc##1{\textcolor[rgb]{0.40,0.40,0.40}{##1}}}
\expandafter\def\csname PYGdefault@tok@gp\endcsname{\let\PYGdefault@bf=\textbf\def\PYGdefault@tc##1{\textcolor[rgb]{0.00,0.00,0.50}{##1}}}
\expandafter\def\csname PYGdefault@tok@o\endcsname{\def\PYGdefault@tc##1{\textcolor[rgb]{0.40,0.40,0.40}{##1}}}
\expandafter\def\csname PYGdefault@tok@kr\endcsname{\let\PYGdefault@bf=\textbf\def\PYGdefault@tc##1{\textcolor[rgb]{0.00,0.50,0.00}{##1}}}
\expandafter\def\csname PYGdefault@tok@s\endcsname{\def\PYGdefault@tc##1{\textcolor[rgb]{0.73,0.13,0.13}{##1}}}
\expandafter\def\csname PYGdefault@tok@kp\endcsname{\def\PYGdefault@tc##1{\textcolor[rgb]{0.00,0.50,0.00}{##1}}}
\expandafter\def\csname PYGdefault@tok@w\endcsname{\def\PYGdefault@tc##1{\textcolor[rgb]{0.73,0.73,0.73}{##1}}}
\expandafter\def\csname PYGdefault@tok@kt\endcsname{\def\PYGdefault@tc##1{\textcolor[rgb]{0.69,0.00,0.25}{##1}}}
\expandafter\def\csname PYGdefault@tok@ow\endcsname{\let\PYGdefault@bf=\textbf\def\PYGdefault@tc##1{\textcolor[rgb]{0.67,0.13,1.00}{##1}}}
\expandafter\def\csname PYGdefault@tok@sb\endcsname{\def\PYGdefault@tc##1{\textcolor[rgb]{0.73,0.13,0.13}{##1}}}
\expandafter\def\csname PYGdefault@tok@k\endcsname{\let\PYGdefault@bf=\textbf\def\PYGdefault@tc##1{\textcolor[rgb]{0.00,0.50,0.00}{##1}}}
\expandafter\def\csname PYGdefault@tok@se\endcsname{\let\PYGdefault@bf=\textbf\def\PYGdefault@tc##1{\textcolor[rgb]{0.73,0.40,0.13}{##1}}}
\expandafter\def\csname PYGdefault@tok@sd\endcsname{\let\PYGdefault@it=\textit\def\PYGdefault@tc##1{\textcolor[rgb]{0.73,0.13,0.13}{##1}}}

\makeatother

\makeatletter
\def\PYG@reset{\let\PYG@it=\relax \let\PYG@bf=\relax%
    \let\PYG@ul=\relax \let\PYG@tc=\relax%
    \let\PYG@bc=\relax \let\PYG@ff=\relax}
\def\PYG@tok#1{\csname PYG@tok@#1\endcsname}
\def\PYG@toks#1+{\ifx\relax#1\empty\else%
    \PYG@tok{#1}\expandafter\PYG@toks\fi}
\def\PYG@do#1{\PYG@bc{\PYG@tc{\PYG@ul{%
    \PYG@it{\PYG@bf{\PYG@ff{#1}}}}}}}
\def\PYG#1#2{\PYG@reset\PYG@toks#1+\relax+\PYG@do{#2}}

\expandafter\def\csname PYG@tok@gd\endcsname{\def\PYG@tc##1{\textcolor[rgb]{0.63,0.00,0.00}{##1}}}
\expandafter\def\csname PYG@tok@gu\endcsname{\let\PYG@bf=\textbf\def\PYG@tc##1{\textcolor[rgb]{0.50,0.00,0.50}{##1}}}
\expandafter\def\csname PYG@tok@gt\endcsname{\def\PYG@tc##1{\textcolor[rgb]{0.00,0.27,0.87}{##1}}}
\expandafter\def\csname PYG@tok@gs\endcsname{\let\PYG@bf=\textbf}
\expandafter\def\csname PYG@tok@gr\endcsname{\def\PYG@tc##1{\textcolor[rgb]{1.00,0.00,0.00}{##1}}}
\expandafter\def\csname PYG@tok@cm\endcsname{\let\PYG@it=\textit\def\PYG@tc##1{\textcolor[rgb]{0.25,0.50,0.50}{##1}}}
\expandafter\def\csname PYG@tok@vg\endcsname{\def\PYG@tc##1{\textcolor[rgb]{0.10,0.09,0.49}{##1}}}
\expandafter\def\csname PYG@tok@m\endcsname{\def\PYG@tc##1{\textcolor[rgb]{0.40,0.40,0.40}{##1}}}
\expandafter\def\csname PYG@tok@mh\endcsname{\def\PYG@tc##1{\textcolor[rgb]{0.40,0.40,0.40}{##1}}}
\expandafter\def\csname PYG@tok@go\endcsname{\def\PYG@tc##1{\textcolor[rgb]{0.53,0.53,0.53}{##1}}}
\expandafter\def\csname PYG@tok@ge\endcsname{\let\PYG@it=\textit}
\expandafter\def\csname PYG@tok@vc\endcsname{\def\PYG@tc##1{\textcolor[rgb]{0.10,0.09,0.49}{##1}}}
\expandafter\def\csname PYG@tok@il\endcsname{\def\PYG@tc##1{\textcolor[rgb]{0.40,0.40,0.40}{##1}}}
\expandafter\def\csname PYG@tok@cs\endcsname{\let\PYG@it=\textit\def\PYG@tc##1{\textcolor[rgb]{0.25,0.50,0.50}{##1}}}
\expandafter\def\csname PYG@tok@cp\endcsname{\def\PYG@tc##1{\textcolor[rgb]{0.74,0.48,0.00}{##1}}}
\expandafter\def\csname PYG@tok@gi\endcsname{\def\PYG@tc##1{\textcolor[rgb]{0.00,0.63,0.00}{##1}}}
\expandafter\def\csname PYG@tok@gh\endcsname{\let\PYG@bf=\textbf\def\PYG@tc##1{\textcolor[rgb]{0.00,0.00,0.50}{##1}}}
\expandafter\def\csname PYG@tok@ni\endcsname{\let\PYG@bf=\textbf\def\PYG@tc##1{\textcolor[rgb]{0.60,0.60,0.60}{##1}}}
\expandafter\def\csname PYG@tok@nl\endcsname{\def\PYG@tc##1{\textcolor[rgb]{0.63,0.63,0.00}{##1}}}
\expandafter\def\csname PYG@tok@nn\endcsname{\let\PYG@bf=\textbf\def\PYG@tc##1{\textcolor[rgb]{0.00,0.00,1.00}{##1}}}
\expandafter\def\csname PYG@tok@no\endcsname{\def\PYG@tc##1{\textcolor[rgb]{0.53,0.00,0.00}{##1}}}
\expandafter\def\csname PYG@tok@na\endcsname{\def\PYG@tc##1{\textcolor[rgb]{0.49,0.56,0.16}{##1}}}
\expandafter\def\csname PYG@tok@nb\endcsname{\def\PYG@tc##1{\textcolor[rgb]{0.00,0.50,0.00}{##1}}}
\expandafter\def\csname PYG@tok@nc\endcsname{\let\PYG@bf=\textbf\def\PYG@tc##1{\textcolor[rgb]{0.00,0.00,1.00}{##1}}}
\expandafter\def\csname PYG@tok@nd\endcsname{\def\PYG@tc##1{\textcolor[rgb]{0.67,0.13,1.00}{##1}}}
\expandafter\def\csname PYG@tok@ne\endcsname{\let\PYG@bf=\textbf\def\PYG@tc##1{\textcolor[rgb]{0.82,0.25,0.23}{##1}}}
\expandafter\def\csname PYG@tok@nf\endcsname{\def\PYG@tc##1{\textcolor[rgb]{0.00,0.00,1.00}{##1}}}
\expandafter\def\csname PYG@tok@si\endcsname{\let\PYG@bf=\textbf\def\PYG@tc##1{\textcolor[rgb]{0.73,0.40,0.53}{##1}}}
\expandafter\def\csname PYG@tok@s2\endcsname{\def\PYG@tc##1{\textcolor[rgb]{0.73,0.13,0.13}{##1}}}
\expandafter\def\csname PYG@tok@vi\endcsname{\def\PYG@tc##1{\textcolor[rgb]{0.10,0.09,0.49}{##1}}}
\expandafter\def\csname PYG@tok@nt\endcsname{\let\PYG@bf=\textbf\def\PYG@tc##1{\textcolor[rgb]{0.00,0.50,0.00}{##1}}}
\expandafter\def\csname PYG@tok@nv\endcsname{\def\PYG@tc##1{\textcolor[rgb]{0.10,0.09,0.49}{##1}}}
\expandafter\def\csname PYG@tok@s1\endcsname{\def\PYG@tc##1{\textcolor[rgb]{0.73,0.13,0.13}{##1}}}
\expandafter\def\csname PYG@tok@sh\endcsname{\def\PYG@tc##1{\textcolor[rgb]{0.73,0.13,0.13}{##1}}}
\expandafter\def\csname PYG@tok@sc\endcsname{\def\PYG@tc##1{\textcolor[rgb]{0.73,0.13,0.13}{##1}}}
\expandafter\def\csname PYG@tok@sx\endcsname{\def\PYG@tc##1{\textcolor[rgb]{0.00,0.50,0.00}{##1}}}
\expandafter\def\csname PYG@tok@bp\endcsname{\def\PYG@tc##1{\textcolor[rgb]{0.00,0.50,0.00}{##1}}}
\expandafter\def\csname PYG@tok@c1\endcsname{\let\PYG@it=\textit\def\PYG@tc##1{\textcolor[rgb]{0.25,0.50,0.50}{##1}}}
\expandafter\def\csname PYG@tok@kc\endcsname{\let\PYG@bf=\textbf\def\PYG@tc##1{\textcolor[rgb]{0.00,0.50,0.00}{##1}}}
\expandafter\def\csname PYG@tok@c\endcsname{\let\PYG@it=\textit\def\PYG@tc##1{\textcolor[rgb]{0.25,0.50,0.50}{##1}}}
\expandafter\def\csname PYG@tok@mf\endcsname{\def\PYG@tc##1{\textcolor[rgb]{0.40,0.40,0.40}{##1}}}
\expandafter\def\csname PYG@tok@err\endcsname{\def\PYG@bc##1{\setlength{\fboxsep}{0pt}\fcolorbox[rgb]{1.00,0.00,0.00}{1,1,1}{\strut ##1}}}
\expandafter\def\csname PYG@tok@kd\endcsname{\let\PYG@bf=\textbf\def\PYG@tc##1{\textcolor[rgb]{0.00,0.50,0.00}{##1}}}
\expandafter\def\csname PYG@tok@ss\endcsname{\def\PYG@tc##1{\textcolor[rgb]{0.10,0.09,0.49}{##1}}}
\expandafter\def\csname PYG@tok@sr\endcsname{\def\PYG@tc##1{\textcolor[rgb]{0.73,0.40,0.53}{##1}}}
\expandafter\def\csname PYG@tok@mo\endcsname{\def\PYG@tc##1{\textcolor[rgb]{0.40,0.40,0.40}{##1}}}
\expandafter\def\csname PYG@tok@kn\endcsname{\let\PYG@bf=\textbf\def\PYG@tc##1{\textcolor[rgb]{0.00,0.50,0.00}{##1}}}
\expandafter\def\csname PYG@tok@mi\endcsname{\def\PYG@tc##1{\textcolor[rgb]{0.40,0.40,0.40}{##1}}}
\expandafter\def\csname PYG@tok@gp\endcsname{\let\PYG@bf=\textbf\def\PYG@tc##1{\textcolor[rgb]{0.00,0.00,0.50}{##1}}}
\expandafter\def\csname PYG@tok@o\endcsname{\def\PYG@tc##1{\textcolor[rgb]{0.40,0.40,0.40}{##1}}}
\expandafter\def\csname PYG@tok@kr\endcsname{\let\PYG@bf=\textbf\def\PYG@tc##1{\textcolor[rgb]{0.00,0.50,0.00}{##1}}}
\expandafter\def\csname PYG@tok@s\endcsname{\def\PYG@tc##1{\textcolor[rgb]{0.73,0.13,0.13}{##1}}}
\expandafter\def\csname PYG@tok@kp\endcsname{\def\PYG@tc##1{\textcolor[rgb]{0.00,0.50,0.00}{##1}}}
\expandafter\def\csname PYG@tok@w\endcsname{\def\PYG@tc##1{\textcolor[rgb]{0.73,0.73,0.73}{##1}}}
\expandafter\def\csname PYG@tok@kt\endcsname{\def\PYG@tc##1{\textcolor[rgb]{0.69,0.00,0.25}{##1}}}
\expandafter\def\csname PYG@tok@ow\endcsname{\let\PYG@bf=\textbf\def\PYG@tc##1{\textcolor[rgb]{0.67,0.13,1.00}{##1}}}
\expandafter\def\csname PYG@tok@sb\endcsname{\def\PYG@tc##1{\textcolor[rgb]{0.73,0.13,0.13}{##1}}}
\expandafter\def\csname PYG@tok@k\endcsname{\let\PYG@bf=\textbf\def\PYG@tc##1{\textcolor[rgb]{0.00,0.50,0.00}{##1}}}
\expandafter\def\csname PYG@tok@se\endcsname{\let\PYG@bf=\textbf\def\PYG@tc##1{\textcolor[rgb]{0.73,0.40,0.13}{##1}}}
\expandafter\def\csname PYG@tok@sd\endcsname{\let\PYG@it=\textit\def\PYG@tc##1{\textcolor[rgb]{0.73,0.13,0.13}{##1}}}

\makeatother
\usepackage[bookmarks=false]{hyperref}
\begin{document}
\title{SqueezeSegV3: Spatially-Adaptive Convolution for Efficient Point-Cloud Segmentation} 

\titlerunning{SqueezeSegV3}
%
\author{Chenfeng Xu\inst{1} \and
Bichen Wu\inst{2} \and Zining Wang\inst{1}  \and Wei Zhan\inst{1} \and Peter Vajda\inst{2} \and Kurt Keutzer\inst{1} \and Masayoshi Tomizuka\inst{1}}
%
\authorrunning{C. Xu et al.}
%

\institute{University of California, Berkeley\\
 \and
 Facebook Inc \\
\email{\{xuchenfeng,wangzining,wzhan,keutzer\}@berkeley.edu},
\email{\{wbc, vajdap\}@fb.com}, \\
 \email{tomizuka@me.berkeley.edu}\\
}
\maketitle    
%
\begin{abstract}
LiDAR point-cloud segmentation is an important problem for many applications. For large-scale point cloud segmentation, the \textit{de facto} method is to project a 3D point cloud to get a 2D LiDAR image and use convolutions to process it. Despite the similarity between regular RGB and LiDAR images, we discover that the feature distribution of LiDAR images changes drastically at different image locations. Using standard convolutions to process such LiDAR images is problematic, as convolution filters pick up local features that are only active in specific regions in the image. As a result, the capacity of the network is under-utilized and the segmentation performance decreases. To fix this, we propose Spatially-Adaptive Convolution (SAC) to adopt different filters for different locations according to the input image. SAC can be computed efficiently since it can be implemented as a series of element-wise multiplications, im2col, and standard convolution. It is a general framework such that several previous methods can be seen as special cases of SAC. Using SAC, we build SqueezeSegV3 for LiDAR point-cloud segmentation and outperform all previous published methods by at least 3.7\% mIoU on the SemanticKITTI benchmark with comparable inference speed. Code and pretrained model are avalibale at \url{https://github.com/chenfengxu714/SqueezeSegV3}.
\end{abstract}
\section{Introduction}
\label{intro}
LiDAR sensors are widely used in many applications  \cite{xie2019review}, especially autonomous driving \cite{geiger2013vision,wu2017squeezeseg,behley2019iccv}. For level 4 \& 5 autonomous vehicles, most of the solutions rely on LiDAR to obtain a point-cloud representation of the environment. LiDAR point clouds can be used in many ways to understand the environment, such as 2D/3D object detection~\cite{zhou2018voxelnet,chen2017multi,song2016deep,qi2018frustum}, multi-modal fusion \cite{zhou2019end,jaritz2019xmuda}, simultaneous localization and mapping \cite{chen2019suma++,behley2018efficient} and point-cloud segmentation \cite{wu2017squeezeseg,wu2018squeezesegv2,qi2017pointnet}. This paper is focused on point-cloud segmentation. This task takes a point-cloud as input and aims to assign each point a label corresponding to its object category. For autonomous driving, point-cloud segmentation can be used to recognize objects such as pedestrians and cars, identify drivable areas, detecting lanes, and so on. More applications of point-cloud segmentation are discussed in \cite{xie2019review}. 

\begin{figure*}[ht]
\centering
\includegraphics[width=0.8\paperwidth]{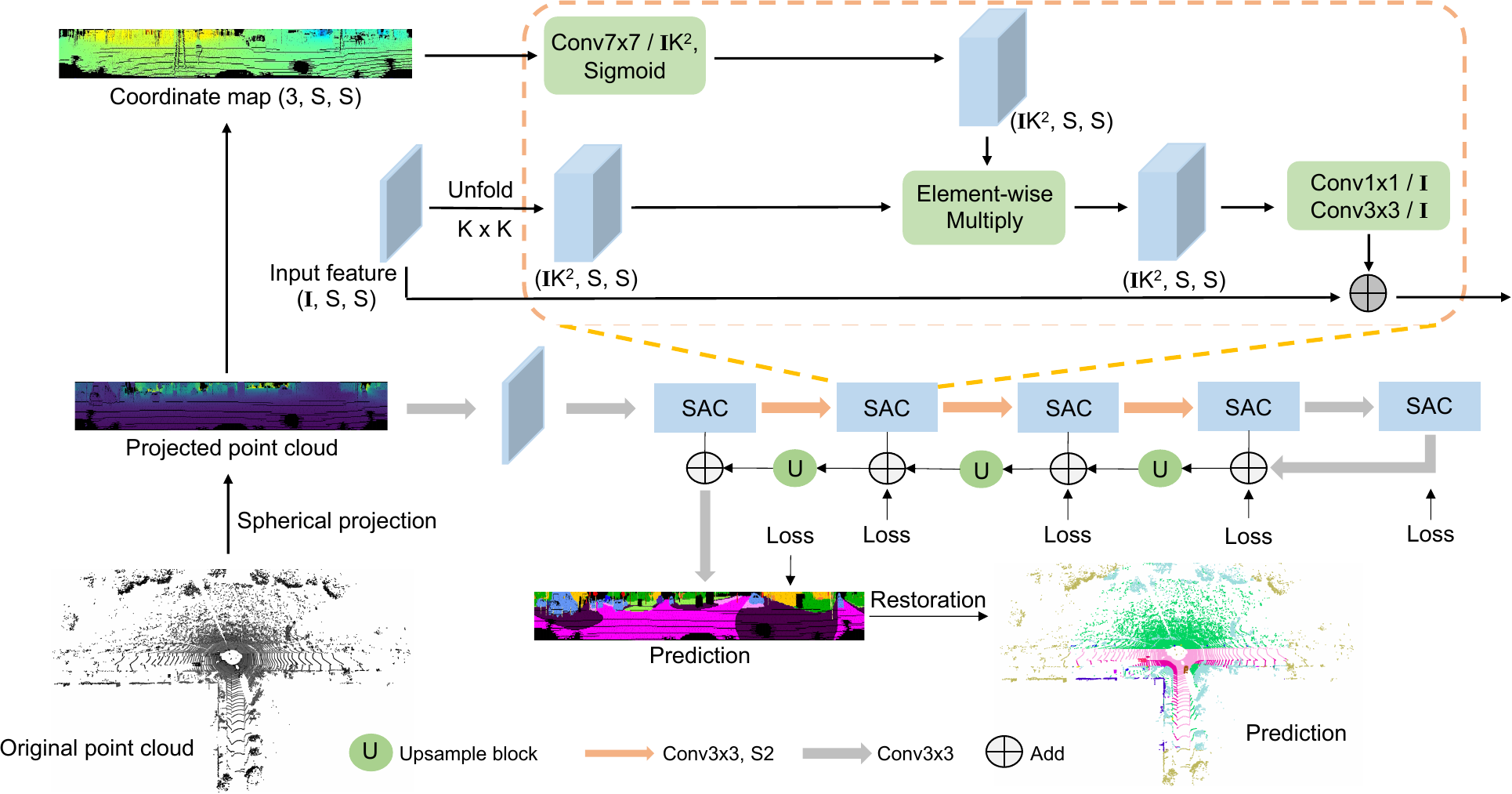}
\caption{The framework of SqueezeSegV3. A LiDAR point cloud is projected to generate a LiDAR image, which is then processed by spatially adaptive convolutions (SAC). The network outputs a point-wise prediction that can be restored to label the 3D point cloud. 
Other variants of SAC can be found in Figure \ref{fig:form}.
}
\label{fig:framework}
\end{figure*}

Recent work on point-cloud segmentation is mainly divided into two categories, focusing on small-scale or large-scale point-clouds. For small-scale problems, ranging from object parsing to indoor scene understanding, most of the recent methods are based on PointNet \cite{qi2017pointnet,qi2017pointnet++}. Although PointNet-based methods have achieved competitive performance in many 3D tasks, they have limited processing speed, especially for large-scale point clouds. For outdoor scenes and applications such as autonomous driving, typical LiDAR sensors, such as Velodyne HDL-64E LiDAR, can scan about $64 \times 3000 =192,000$ points for each frame, covering an area of $160 \times 160 \times 20$ meters. Processing point clouds at such scale efficiently or even in real time is far beyond the capability of PointNet-based methods. Hence, much of the recent work follows the method based on spherical projection proposed by Wu~\textit{et al.} \cite{wu2017squeezeseg,wu2018squeezesegv2}. Instead of processing 3D points directly, these methods first transform a 3D LiDAR point cloud into a 2D LiDAR image and use 2D ConvNets to segment the point cloud, as shown in Figure \ref{fig:framework}. In this paper, we follow this method based on spherical projection.

\vspace*{-6mm}
To transform a 3D point-cloud into a 2D grid representation, each point in the 3D space is projected to a spherical surface. The projection angles of each point are quantized and used to denote the location of the pixel. Each point's original 3D coordinates are treated as features. Such representations of LiDAR are very similar to RGB images, therefore, it seems straightforward to adopt 2D convolution to process ``LiDAR images''. This pipeline is illustrated in Figure \ref{fig:framework}.  

\begin{figure*}[t!]
\centering
\includegraphics[width=0.8\paperwidth]{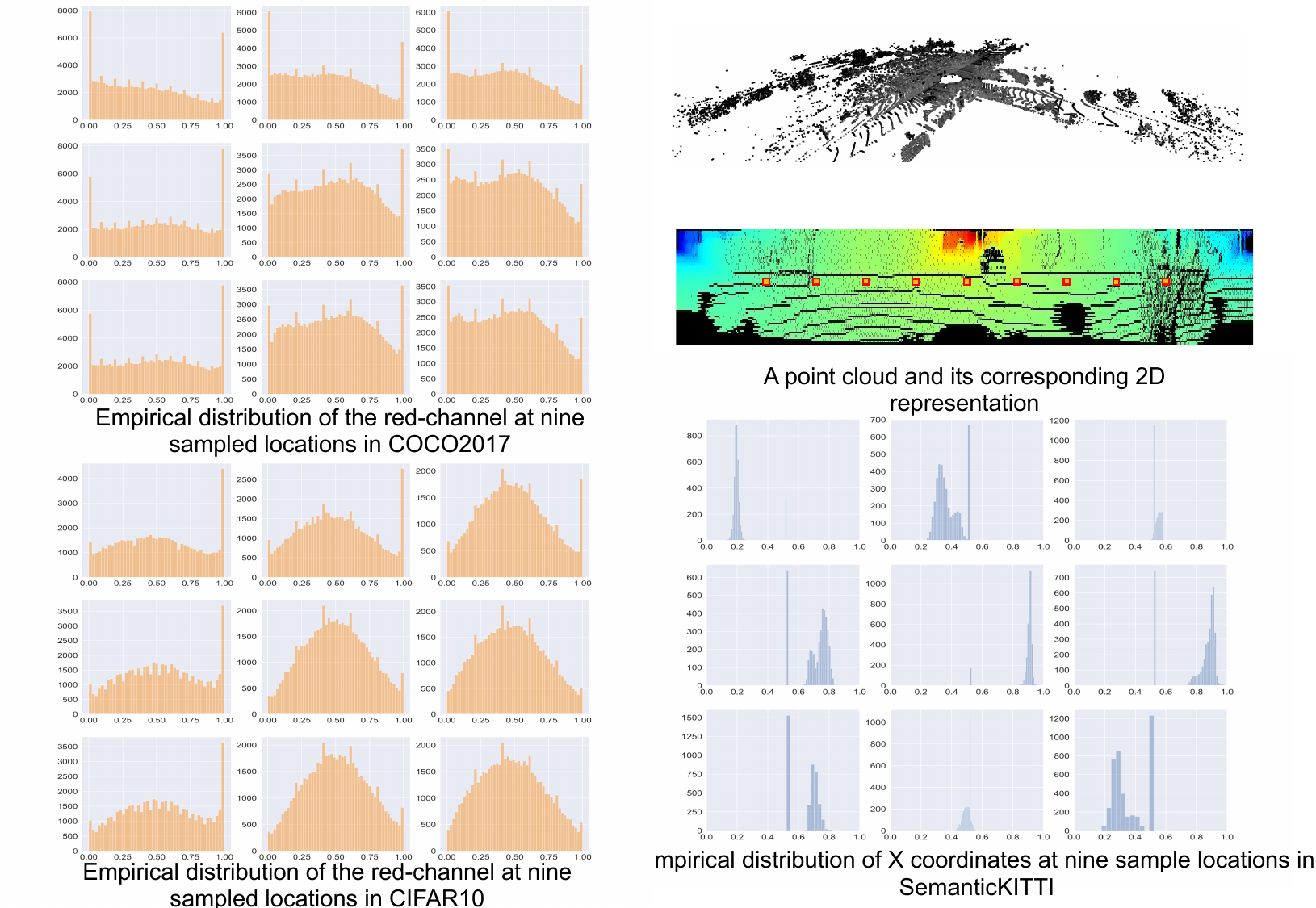}
\caption{Pixel-wise feature distribution at nine sampled locations from COCO2017~\cite{lin2014microsoft}, CIFAR10~\cite{krizhevsky2009learning} and SemanticKITTI~\cite{behley2019iccv}. The left shows the distribution of the red channel across all images in COCO2017 and CIFAR10. The right shows the distribution of the X coordinates across all LiDAR images in SemanticKITTI.}
\label{fig:cifar-vs-kitti}
\centering
\end{figure*}

However, we discovered that an important difference exists between LiDAR images and regular images. For a regular image, the feature distribution is largely invariant to spatial locations, as visualized in Figure \ref{fig:cifar-vs-kitti}. For a LiDAR image, its features are converted by spherical projection, which introduces very strong spatial priors. As a result, the feature distribution of LiDAR images varies drastically at different locations, as illustrated in Figure \ref{fig:cifar-vs-kitti} and Figure \ref{fig:kitti-activation} (top). When we train a ConvNet to process LiDAR images, convolution filters may fit local features and become only active in some regions and are not used in other parts, as confirmed in Figure \ref{fig:kitti-activation} (bottom). As a result, the capacity of the model is under-utilized, leading to decreased performance in point-cloud segmentation. 

To tackle this problem, we propose Spatially-Adaptive Convolution (SAC), as shown in Figure \ref{fig:framework}. SAC is designed to be spatially-adaptive and content-aware. Based on the input, it adapts its filters to process different parts of the image. To ensure efficiency, we factorize the adaptive filter into a product of a static convolution weight and an attention map. The attention map is computed by a one-layer convolution, whose output at each pixel location is used to adapt the static weight. By carefully scheduling the computation, SAC can be implemented as a series of widely supported and optimized operations including element-wise multiplication, im2col, and reshaping, which ensures the efficiency of SAC.

SAC is formulated as a general framework such that previous methods such as squeeze-and-excitation (SE) \cite{hu2018squeeze}, convolutional block attention module (CBAM) \cite{woo2018cbam}, context-aggregation module (CAM) \cite{wu2018squeezesegv2}, and pixel-adaptive convolution (PAC) \cite{su2019pixel} can be seen as special cases of SAC, and experiments show that the more general SAC variants proposed in this paper outperform previous ones.

Using spatially-adaptive convolution, we build SqueezeSegV3 for LiDAR point-cloud segmentation. On the SemanticKITTI benchmark, SqueezeSegV3 outperforms all previously published methods by at least 3.7 mIoU with comparable inference speed, demonstrating the effectiveness of spatially-adaptive convolution.

\section{Related work}
\subsection{Point-Cloud Segmentation}
Recent papers on point-cloud segmentation can be divided into two categories - those that deal with small-scale point-clouds, and those that deal with large-scale point clouds. For small-scale point-cloud segmentation such as object part parsing and indoor scene understanding, mainstream methods are based on PointNet \cite{qi2017pointnet,qi2017pointnet++}. DGCNN \cite{wang2019dynamic} and Deep-KdNet \cite{klokov2017escape} extend the hierarchical architecture of PointNet++~\cite{qi2017pointnet++} by grouping neighbor points. Based on the PointNet architecture, ~\cite{dovrat2019learning,li2018so,li2018pointcnn} further improve the effectiveness of sampling, reordering and grouping to obtain a better representation for downstream tasks. PVCNN~\cite{liu2019point} improves the efficiency of PointNet-based methods~\cite{liu2019point,wang2019dynamic} using voxel-based convolution with a contiguous memory access pattern. Despite these efforts, the efficiency of PointNet-based methods is still limited since they inherently need to process sparse data, which is more difficult to accelerate~\cite{liu2019point}. It is noteworthy to mention that the most recent RandLA-Net~\cite{hu2019randla} significantly improves the speed of point cloud processing in the novel use of random sampling. 

Large-scale point-cloud segmentation is challenging since 1) large-scale point-clouds are difficult to annotate and 2) many applications require real-time inference. Since a typical outdoor LiDAR (such as Velodyne HDL-64E) can collect about $200K$ points per scan, it is difficult for previous methods \cite{landrieu2018large,riegler2017octnet,tchapmi2017segcloud,liu20173dcnn,rethage2018fully,mo2019partnet,meng2019vv} to satisfy a real-time latency constraint. To address the data challenge, \cite{wu2017squeezeseg,wang2019latte} proposed tools to label 3D bounding boxes and convert to point-wise segmentation labels. \cite{wu2017squeezeseg,wu2018squeezesegv2,yue2018lidar} proposed to train with simulated data. Recently, Behley~\textit{et al.} proposed SemanticKITTI \cite{behley2019iccv}, a densely annotated dataset for large-scale point-cloud segmentation. For efficiency, Wu~\textit{et al.} \cite{wu2017squeezeseg} proposed to project 3D point clouds to 2D and transform point-cloud segmentation to image segmentation. Later work \cite{wu2018squeezesegv2,behley2019iccv,milioto2019rangenet++} continued to improve the projection-based method, making it a popular choice for large-scale point-cloud segmentation. 

\subsection{Adaptive Convolution}
Standard convolutions use the same weights to process input features at all spatial locations regardless of the input. Adaptive convolutions may change the weights according to the input and the location in the image. Squeeze-and-excitation and its variants \cite{hu2018squeeze,hu2018gather,woo2018cbam} compute channel-wise or spatial attention to adapt the output feature map. Pixel-adaptive convolution (PAC) \cite{su2019pixel} changes the convolution weight along the kernel dimension with a Gaussian function. Wang~\textit{et al.}~\cite{wang2018depth} propose to directly re-weight the standard convolution with a depth-aware Gaussian kernel. 3DNConv~\cite{chen20193d} further extends \cite{wang2018depth} by estimating depth through an RGB image and using it to improve image segmentation. In our work, we propose a more general framework such that channel-wise attention \cite{hu2018squeeze,hu2018gather}, spatial attention \cite{woo2018cbam,wu2018squeezesegv2} and PAC \cite{su2019pixel} can be considered as special cases of spatially-adaptive convolution. 
In addition to adapting weights, deformable convolutions \cite{dai2017deformable,zhu2019deformable} adapt the location to pull features to convolution. DKN \cite{kim2019deformable} combines both deformable convolution and adaptive convolution for joint-image filtering. However, deformable convolution is orthogonal to our proposed method.

\subsection{Efficient Neural Networks}
Many applications that involve point-cloud segmentation require real-time inference. To meet this requirement, we not only need to design efficient segmentation pipelines \cite{wu2018squeezesegv2}, but also efficient neural networks which optimize the parameter size, FLOPs, latency, power, and so on \cite{wu2019efficient}. 

Many  neural nets have been targeted to achieve efficiency, including SqueezeNet \cite{iandola2016squeezenet,gholami2018squeezenext,wu2017squeezedet}, MobileNets \cite{howard2017mobilenets,sandler2018mobilenetv2,howard2019searching},  ShiftNet \cite{wu2018shift,yang2019synetgy}, ShuffleNet \cite{zhang2018shufflenet,ma2018shufflenet}, FBNet \cite{wu2019fbnet,wu2018mixed}, ChamNet \cite{dai2019chamnet}, MnasNet \cite{tan2019mnasnet}, and EfficientNet \cite{tan2019efficientnet}. Previous work shows that using a more efficient backbone network can effectively improve efficiency in downstream tasks. In this paper, however, in order to rigorously evaluate the performance of spatially-adaptive convolution (SAC), we use the same backbone as RangeNet++ \cite{milioto2019rangenet++}.

\section{Spherical Projection of LiDAR Point-Cloud}
To process a LiDAR point-cloud efficiently, Wu~\textit{et al.} \cite{wu2017squeezeseg} proposed a pipeline (shown in Figure \ref{fig:framework}) to project a sparse 3D point cloud to a 2D LiDAR image as
\begin{equation}
[\begin{array}{c}
     p \\
     q
\end{array}] = [\begin{array}{c}
    \frac{1}{2}(1 - arctan(y,x) / \pi) \cdot w\\
    (1 - (arcsin(z\cdot r^{-1})+f_{up})\cdot f^{-1}) \cdot h
\end{array}],
\label{eqn:spherical}
\end{equation}
where $(x, y, z)$ are 3D coordinates, $(p, q)$ are angular coordinates, $(h, w)$ are the height and width of the desired projected 2D map, $f = f_{up} + f_{down}$ is the vertical field-of-view of the LiDAR sensor, and $r =\sqrt{x^2+y^2+z^2} $ is the range of each point. For each point projected to $(p, q)$, we use its measurement of $(x, y, z, r)$ and intensity as features and stack them along the channel dimension. This way, we can represent a LiDAR point cloud as a LiDAR image with the shape of $(h, w, 5)$. Point-cloud segmentation can then be reduced to image segmentation, which is typically solved using ConvNets.

Despite the apparent similarity between LiDAR and RGB images, we discover that the spatial distribution of RGB features are quite different from $(x, y, z, r)$ features. In Figure \ref{fig:cifar-vs-kitti}, we sample nine pixels on images from COCO~\cite{lin2014microsoft}, CIFAR10 \cite{krizhevsky2009learning} and SemanticKITTI \cite{behley2019iccv} and compare their feature distribution. In COCO and CIFAR10, the feature distribution at different locations are rather similar. For SemanticKITTI, however, feature distribution at each locations are drastically different. Such spatially-varying distribution is caused by the spherical projection in Equation (\ref{eqn:spherical}). In Figure \ref{fig:kitti-activation} (top), we plot the mean of x, y, and z channels of LiDAR images. Along the width dimension, we can see the sinusoidal change of x and y channels. Along the height dimension, points projected to the top of the image have higher z-values than the ones projected to the bottom. As we will discuss later, such spatially varying distribution can degrade the performance of convolutions.

\section{Spatially-Adaptive Convolution}
\subsection{Standard Convolution}
Previous methods based on spherical projection \cite{wu2017squeezeseg,wu2018squeezesegv2,milioto2019rangenet++} treat projected LiDAR images as RGB images and process them with standard convolution as
\begin{equation}
    Y[m, p, q] = \sigma(\sum_{i,j,n} W[m,n,i,j] \times X[n,p + \hat{i}, q+\hat{j}]),
    \label{eqn:conv}
\end{equation}
where $Y \in \mathbf{R}^{O \times S \times S}$ is the output tensor, $X \in \mathbf{R}^{I\times S \times S}$ denotes the input tensor, and $W \in \mathbf{R}^{O\times I\times K\times K}$ is the convolution weight. $O, I, S, K$ are the output channel size, input channel size, image size, and kernel size of the weight, respectively. $\hat{i} = i - \left \lfloor{K / 2}\right \rfloor $, $\hat{j} = j - \left \lfloor{K / 2}\right \rfloor.$ $\sigma(\cdot)$ is a non-linear activation function. 

Convolution is based on a strong inductive bias that the distribution of visual features is invariant to image locations. For RGB images, this is a somewhat valid assumption, as illustrated in Figure \ref{fig:cifar-vs-kitti}. Therefore, regardless of the location, a convolution use the same weight $W$ to process the input. This design makes the convolution operation very computationally efficient: First, convolutional layers are efficient in parameter size. Regardless of the input resolution $S$, a convolutional layer's parameter size remains the same as $O \times I \times K \times K$. Second, convolution is efficient to compute. In modern computer architectures, loading parameters into memory costs orders-of-magnitude higher energy and latency than floating point operations such as multiplications and additions 
\cite{pedram2016dark}. 
For convolutions, we can load the parameter once and re-use for all the input pixels, which significantly improves the latency and power efficiency. 

\begin{figure*}[ht]
\centering
\includegraphics[width=0.8\paperwidth]{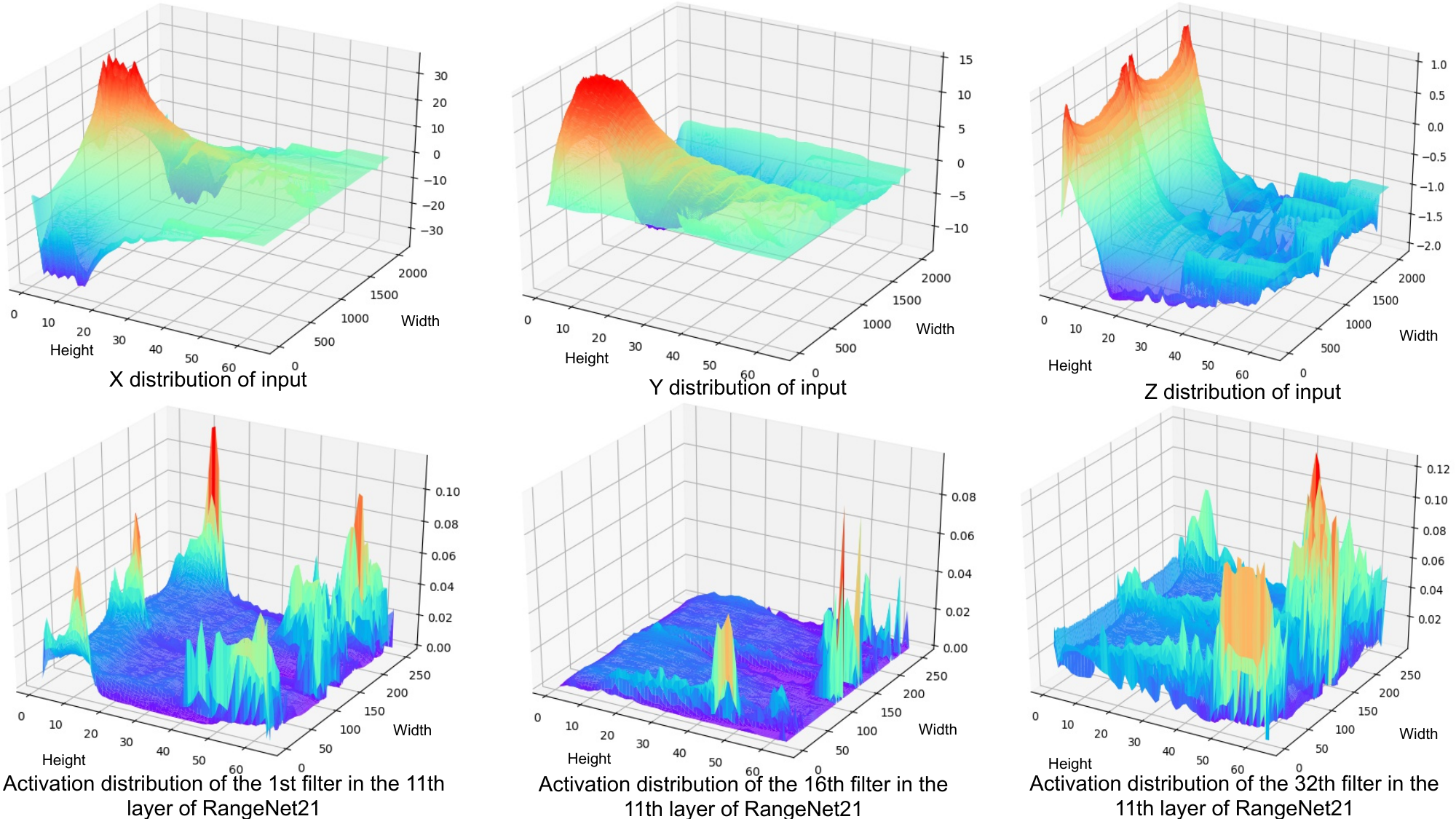}
\caption{Channel and filter activation visualization on the SemanticKITTI dataset. Top: we visualize the mean value of x, y, and z channels of the projected LiDAR images at different locations. Along the width dimension, we can see the sinusoidal change of the x and y channels. Along the height dimension, we can see z values are higher at the top of the image. Bottom: We visualize the mean activation value of three filters at the 11th layer of a pre-trained RangeNet21 \cite{milioto2019rangenet++}. We can see that those filters are sparsely activated only in certain areas.}
\label{fig:kitti-activation}
\centering
\end{figure*}

\begin{figure*}[t!]

\centering
\begin{subfigure}[b]{0.75\textwidth}
         \centering
         \includegraphics[width=\textwidth]{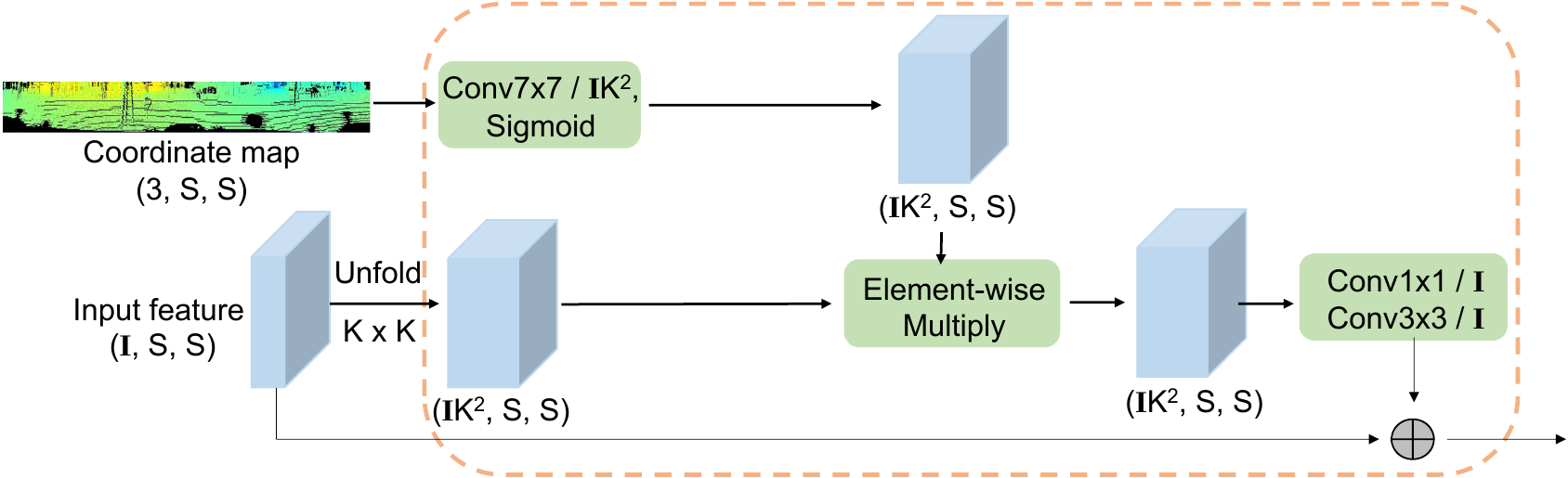}
         \caption{SAC-ISK}
         \hrule
         \label{fig:SAC-ISK}
     \end{subfigure}
     \hfill
     \begin{subfigure}[b]{0.75\textwidth}
         \centering
         \includegraphics[width=\textwidth]{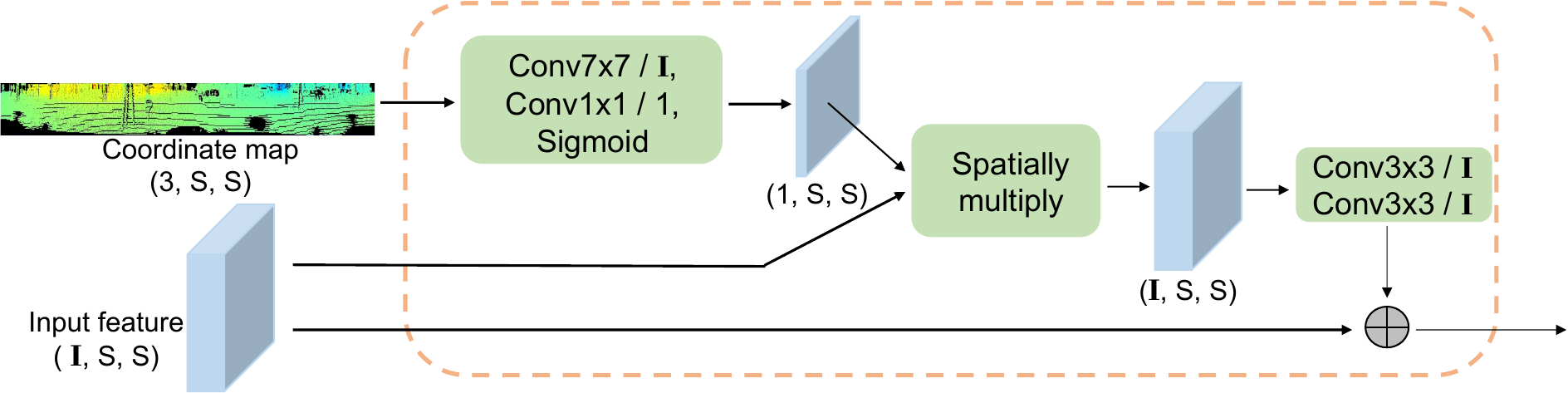}
         \caption{SAC-S}
         \hrule
         \label{fig:SAC-S}
     \end{subfigure}
     \hfill
     \begin{subfigure}[b]{0.75\textwidth}
         \centering
         \includegraphics[width=\textwidth]{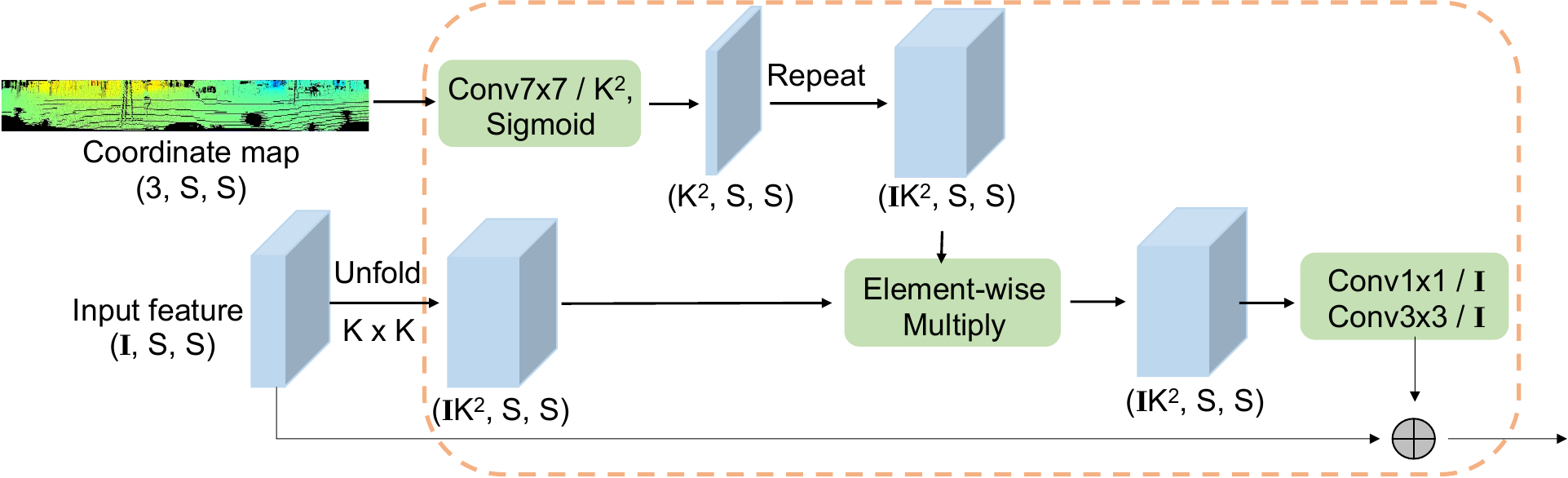}
         \caption{SAC-SK}
         \hrule
         \label{fig:SAC-SK}
     \end{subfigure}
     \hfill
     \begin{subfigure}[b]{0.75\textwidth}
         \centering
         \includegraphics[width=\textwidth]{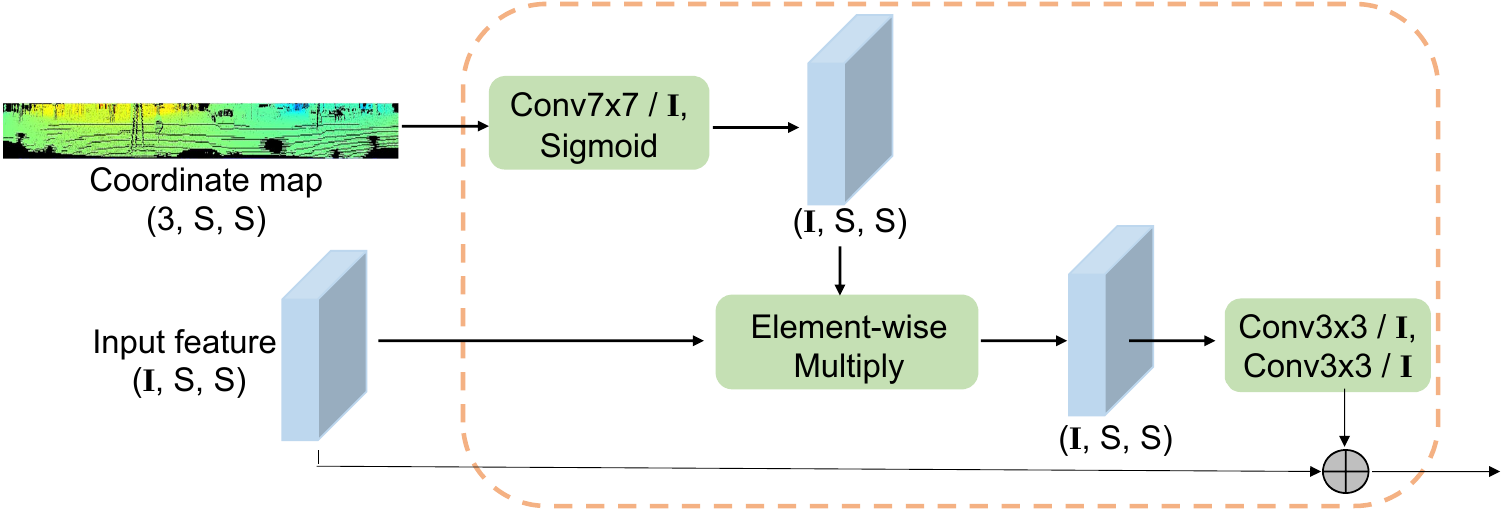}
         \caption{SAC-IS}
         \label{fig:SAC-IS}
     \end{subfigure}
          \hfill
        \caption{Variants of spatially-adaptive convolution used in Figure \ref{fig:framework}.}
        \label{fig:form}
\centering
\end{figure*}

However, for LiDAR images, the feature distribution across the image are no longer identical, as illustrated in Figure \ref{fig:cifar-vs-kitti} and \ref{fig:kitti-activation} (top). Many features may only exist in local regions of the image, so the filters that are trained to process them are only active in the corresponding regions and are not useful elsewhere. To confirm this, we analyze a trained RangeNet21 \cite{milioto2019rangenet++} by calculating the average filter activation across the image. We can see in Figure \ref{fig:kitti-activation} (bottom) that convolutional filters are sparsely activated and remain zero in many regions. This validates that convolution filters are spatially under-utilized. 
\subsection{Spatially-Adaptive Convolution}
\label{sec:sac-variant}
To better process LiDAR images with spatially-varying feature distributions, we re-design convolution to achieve two goals: 1) It should be spatially-adaptive and content-aware. The new operator should process different parts of the image with different filters, and the filters should adapt to feature variations. 2) The new operator should be efficient to compute. 

To achieve these goals, we propose Spatially-Adaptive Convolution (SAC), which can be described as the following:
 \begin{equation}
    Y[m, p, q] = \sigma(\sum_{i,j,n}  W(X_0)[m,n,p,q,i,j] \times  X[n,p + \hat{i} , q+\hat{j}]) .
    \label{eqn:sac}
\end{equation}
$W(\cdot) \in \mathbf{R}^{O\times I\times S\times S\times K\times K}$ is a function of the raw input $X_0$.  It is spatially-adaptive, since $W$ depends on the location $(p, q)$. It is content-aware since $W$ is a function of the raw input $X_0$. Computing $W$ in this general form is very expensive since $W$ contains too many elements to compute.

To reduce the computational cost, we factorize $W$ as the product of a standard convolution weight and a spatially-adaptive attention map as:
\begin{equation}
W[m,n,p,q,i,j] = \hat{W}[m, n, i, j] \times A(X_0)[m, n, p, q, i, j].
\label{eqn:sac-factored}
\end{equation}
$\hat{W} \in \mathbf{R}^{O\times I\times S \times S}$ is a standard convolution weight, and $A \in \mathbf{R}^{O\times I\times S\times S\times K\times K}$ is the attention map. To reduce the complexity, we collapse several dimensions of $A$ to obtain a smaller attention map to make it computationally tractable. 

We denote the first dimension of $A$ as the output channel dimension (O), the second as the input channel dimension (I), the 3rd and 4th dimensions as spatial dimensions (S), and the last two dimensions as kernel dimensions (K). 

Starting from Equation (\ref{eqn:sac-factored}), we name this form of SAC as SAC-OISK, and we re-write $A$ as $A_{OISK}$, where the subscripts denote the dimensions that are not collapsed to 1. If we collapse the output dimension, we name the variant as SAC-ISK, and the attention map as $A_{ISK} \in \mathbf{R}^{1 \times I\times S\times S\times K\times K}$. SAC-ISK adapts a convolution weight spatially as well as across the kernel and input channel dimensions, as shown in Figure \ref{fig:SAC-ISK}. We can further compress the kernel dimensions to obtain SAC-IS with $A_{IS} \in \mathbf{R}^{1 \times I\times S\times S\times 1\times 1}$, (Figure \ref{fig:SAC-IS}) and SAC-S with pixel-wise attention as $A_{S} \in \mathbf{R}^{1 \times 1 \times S\times S\times 1\times 1}$ (Figure \ref{fig:SAC-S}). 


As long as we retain the spatial dimension $A$, SAC is able to spatially adapt a standard convolution. Experiments show that all variants of SAC effectively improve the performance on the SemanticKITTI dataset. 

\subsection{Efficient Computation of SAC}
To efficiently compute an attention map, we feed the raw LiDAR image $X_0$ into a 7x7 convolution followed by a sigmoid activation. The convolution computes the values of the attention map at each location. The more dimensions to adapt, the more FLOPs and parameter size SAC requires. However, most of the variants of SAC are very efficient. Taking SqueezeSegV3-21 as an example, the cost of adding different SAC variants is summarized in Table \ref{tab:SAC-Para}. The extra FLOPs (2.4\% - 24.8\%) and parameters (1.1\% - 14.9\%) needed by SAC is quite small.
    
\begin{table}[!t]
  
\footnotesize
\centering
\small
\begin{tabular}{ c c c c c c c}
 \hline
 Method &  O & I & S & K & Extra Params (\%) & Extra MACs (\%)    \\
 \hline
 \hline

SAC-S &  \xmark & \xmark & \cmark & \xmark &  1.1&  2.4\\
SAC-IS & \xmark & \cmark & \cmark & \xmark &  2.2 & 6.2 \\
SAC-SK & \xmark & \xmark & \cmark & \cmark &  1.9 & 3.1 \\
SAC-ISK & \xmark &\cmark & \cmark & \cmark &  14.9 & 24.8 \\
\hline 
\end{tabular}
\caption{Extra parameters and MACs for different SAC variants in SqueezeSegV3-21}
\vspace{-0.3cm}

\label{tab:SAC-Para}
\end{table}

After obtaining the attention map, we need to efficiently compute the product of the convolution weight $\hat{W}$, attention map $A$, and the input $X$. One choice is to first compute the adaptive weight as Equation (\ref{eqn:sac-factored}) and then process the input $X$. However, the adaptive weight varies per pixel, so we are no longer able to re-use the weight spatially to retain the efficiency of standard convolution. 

So, instead, we first combine the attention map $A$ with the input tensor $X$. For attention maps without kernel dimensions, such as $A_{S}$ or $A_{IS}$, we directly perform element-wise multiplication (with broadcasting) between $A$ and $X$. Then, we apply a standard convolution with weight $W$ on the adapted input. The examples of SAC-S and SAC-IS are illustrated in Figures \ref{fig:SAC-S} and \ref{fig:SAC-IS} respectively. Pseudo-code implementation is provided in the supplementary material. 

For attention maps with kernel dimensions, such as $A_{ISK}$ and $A_{SK}$, we first perform an unfolding (im2col) operation on $X$. At each location, we collect nearby $K$-by-$K$ features and stack them along the channel dimension to get $\tilde{X} \in \mathbf{R}^{K^2I \times S \times S}$. Then, we can apply element-wise multiplication to combine the attention map $A$ and input $X$. Next, we reshape weight $W \in \mathbf{R}^{O\times I\times K\times K}$ as $\tilde{W} \in \mathbf{R}^{O \times K^2I \times 1 \times 1}$. Finally, the output of $Y$ can be obtained by applying a 1-by-1 convolution with $\tilde{W}$ on $\tilde{X}$. The computation of SAC-ISK and SAC-SK is shown in Figures \ref{fig:SAC-ISK} and \ref{fig:SAC-SK} respectively, and the pseudo-code implementation is provided in the supplementary material.

Overall, SAC can be implemented as a series of element-wise multiplications, im2col, reshaping, and standard convolution operations, which are widely supported and well optimized. This ensures that SAC can be computed efficiently.

\begin{figure}[t!]
     \centering
     \begin{subfigure}[b]{0.75\textwidth}
         \centering
         \includegraphics[width=\textwidth]{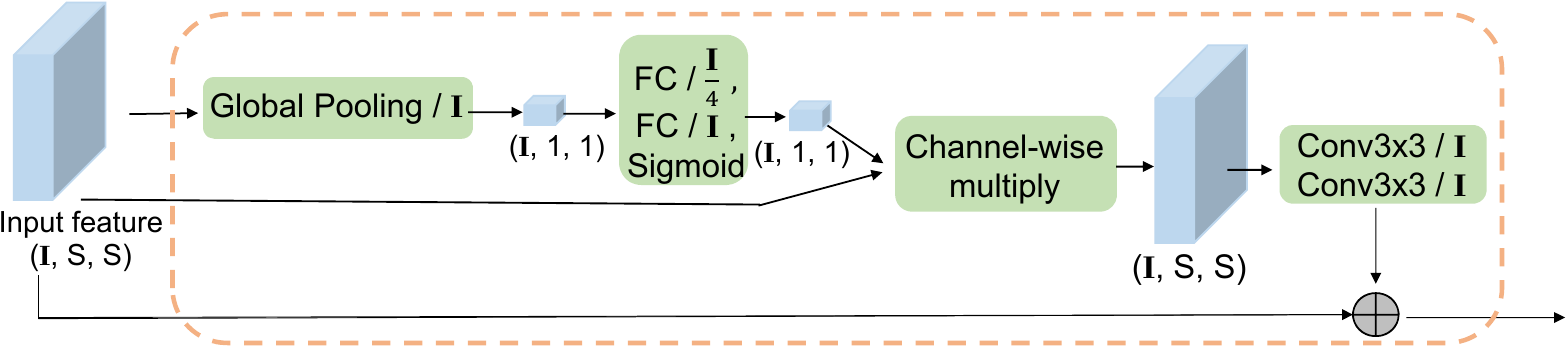}
         \caption{ Squeeze-and-Excitation~\cite{hu2018squeeze}}
         \hrule
         \label{fig:SE}
     \end{subfigure}
     \hfill
          \begin{subfigure}[b]{0.75\textwidth}
         \centering
         \includegraphics[width=\textwidth]{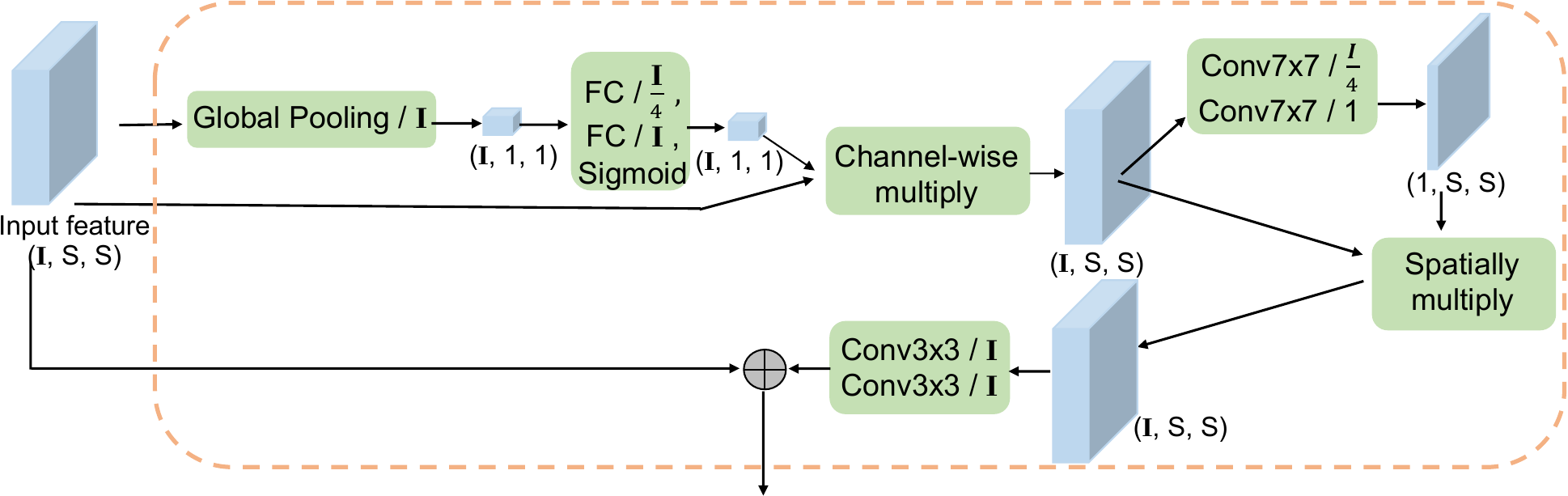}
         \caption{CBAM~\cite{woo2018cbam}.}
         \hrule
         \label{fig:CBAM}
     \end{subfigure}
          \hfill
          \begin{subfigure}[b]{0.75\textwidth}
         \centering
         \includegraphics[width=\textwidth]{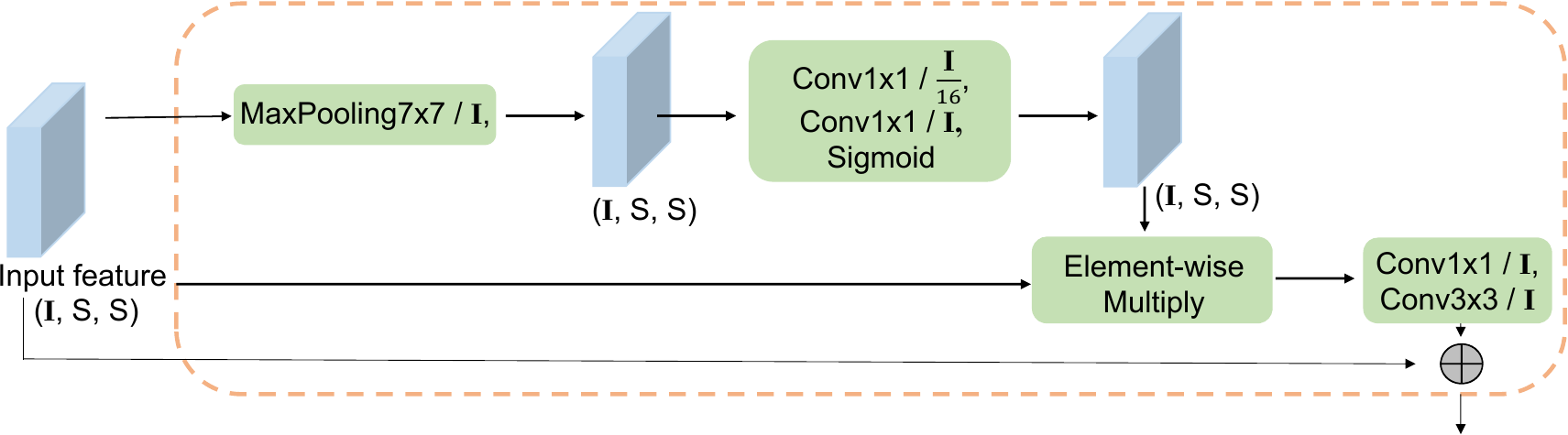}
         \caption{CAM~\cite{wu2018squeezesegv2}}
         \hrule
         \label{fig:CAM}
     \end{subfigure}
               \hfill
          \begin{subfigure}[b]{0.75\textwidth}
         \centering
         \includegraphics[width=\textwidth]{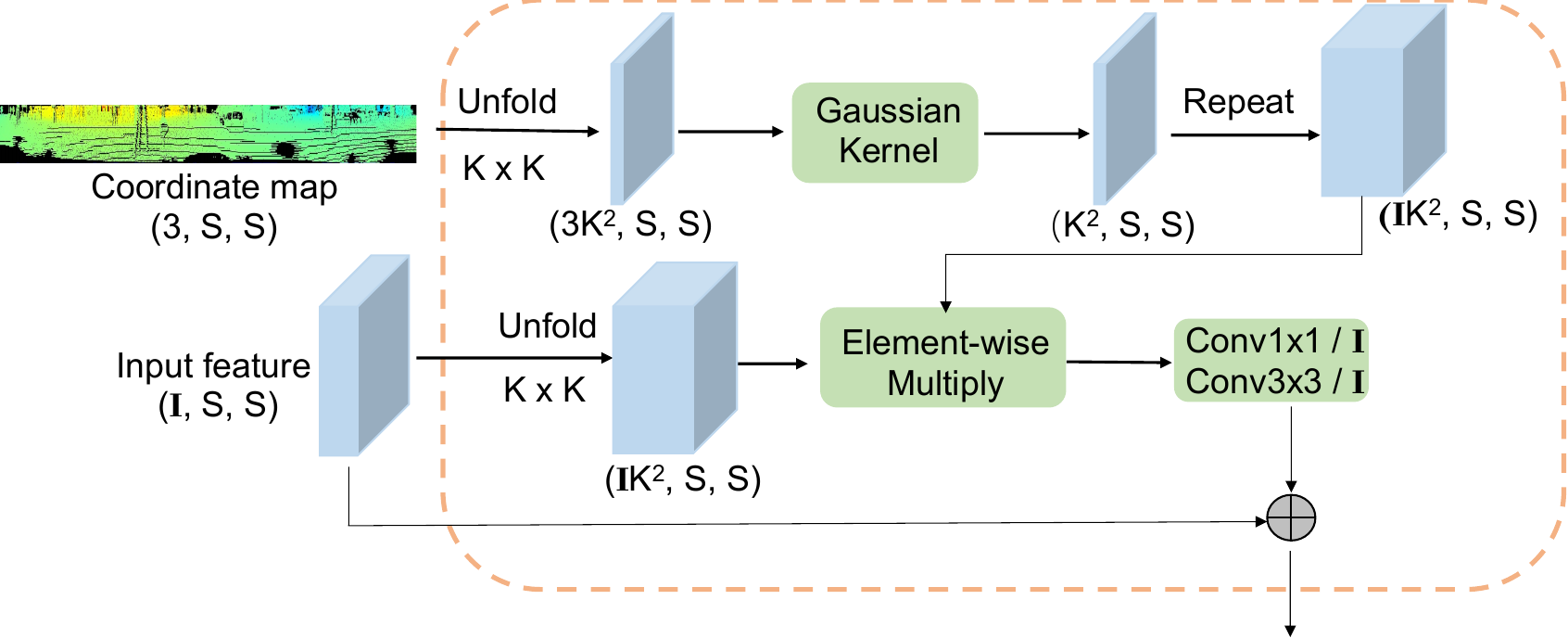}
         \caption{PAC~\cite{su2019pixel}}
         \label{fig:PAC}
     \end{subfigure}
     \hfill
     
    \caption{Variants of spatially-adaptive convolution from previous work.}
    \label{fig:others}
\end{figure}

\subsection{Relationship with Prior Work}
Several prior works can be seen as variants of a spatially-adaptive convolution, as described by Equations \ref{eqn:sac} and \ref{eqn:sac-factored}.
Squeeze-and-Excitation (SE) \cite{hu2018squeeze,hu2018gather} uses global average pooling and fully-connected layers to compute channel-wise attention to adapt the feature map, as illustrated in Figure \ref{fig:SE}. It can be seen as the variant of SAC-I with a attention map of $A_I \in \mathbf{R}^{1\times I \times 1 \times 1 \times 1 \times 1}$. The convolutional block attention module (CBAM) \cite{woo2018cbam} can be see as applying $A_I$ followed by an $A_S$ to adapt the feature map, as shown in Figure \ref{fig:CBAM}. SqueezeSegV2 \cite{wu2018squeezesegv2} uses the context-aggregation module (CAM) to combat dropout noises in LiDAR images. At each position, it uses a 7x7 max pooling followed by 1x1 convolutions to compute a channel-wise attention map. It can be seen as the variant SAC-IS with the attention map of $A_{IS} \in \mathbf{R}^{1\times I\times S\times S\times 1 \times 1}$ as illustrated in Figure \ref{fig:CAM}. Pixel-adaptive convolution (PAC) \cite{su2019pixel} uses a Gaussian function to compute kernel-wise attention for each pixel. It can be seen as the variant of SAC-SK, with the attention map of $A_{SK} \in \mathbf{R}^{1\times 1\times S\times S\times K \times K}$, as illustrated in Figure \ref{fig:PAC}. Our ablation studies compare variants of SAC, including ones proposed in our paper and in prior work. Experiments show our proposed SAC variants outperform previous baselines.

\section{SqueezeSegV3}

Using the spatially-adaptive convolution, we build SqueezeSegV3 for LiDAR point-cloud segmentation. The overview of the model is shown in Figure \ref{fig:framework}. 

\subsection{The Architecture of SqueezeSegV3}
To facilitate rigorous comparison, SqueezeSegV3's backbone architecture is based on RangeNet~\cite{milioto2019rangenet++}. RangeNet contains five stages of convolution, each stage contains several blocks. At the beginning of the stage, it performs downsampling. The output is then upsampled to recover the resolution. Each block of RangeNet contains two stacked convolutions. We replace the first one with SAC-ISK as in Figure \ref{fig:SAC-ISK}. We remove the last two downsampling. To keep the same FLOPs, we reduce the channels of last two stages. The output channel sizes from $Stage1$ to $Stage5$ are 64, 128, 256, 256 and 256 respectively, while the output channel sizes in RangeNet~\cite{milioto2019rangenet++} are 64, 128, 256, 512 and 1024. Due to the removal of the last two downsampling operations, we only adopt 3 upsample blocks using transposed convolution and convolution.

\subsection{Loss Function}

We introduce a multi-layer cross entropy loss to train the proposed network, which is also used in~\cite{shen2017multi,johnson2016perceptual,xu2019learn,newell2016stacked}. During training, from $stage1$ to $stage5$, we add a prediction layer at each stage's output. For each output, we respectively downsample the groundtruth label map by 1x, 2x, 4x, 8x and 8x, and use them to train the output of $stage1$ to $stage5$. The loss function can be described as
\begin{equation}
    L = \sum_{i=1}^5 \frac{- \sum_{H_i, W_i} \sum_{c=1}^C w_c\cdot y_c \cdot log(\hat{y}_c)}{H_i \times W_i}.
\end{equation}
In the equation, $w_c = \frac{1}{log(f_c + \epsilon)}$ is a normalization factor and $f_c$ is the frequency of class $c$. $H_i, W_i$ are the height and width of the output in $i$-th stage, $y_c$ is the prediction for the $c$-th class in each pixel and $\hat{y}_c$ is the label. Compared to the single-stage cross-entropy loss used for the final output, the intermediate supervisions guide the model to form features with more semantic meaning. In addition, they help mitigate the vanishing gradient problem in training.

\section{Experiments}
\subsection{Dataset and Evaluation Metrics}
We conduct our experiments on the SemanticKITTI dataset~\cite{behley2019iccv}, a large-scale dataset for LiDAR point-cloud segmentation. The dataset contains 21 sequences of point-cloud data with 43,442 densely annotated scans and total 4549 millions points. Following \cite{behley2019iccv}, sequences-\{0-7\} and \{9, 10\} (19130 scans) are used for training, sequence-08 (4071 scans) is for validation, and sequences-\{11-21\} (20351 scans) are for test. Following previous work~\cite{milioto2019rangenet++}, we use mIoU over 19 categories to evaluate the accuracy.


\subsection{Implementation Details}
We pre-process all the points by spherical projection following Equation (\ref{eqn:spherical}). The 2D LiDAR images are then processed by SqueezeSegV3 to get a 2D predicted label map, which is then restored back to the 3D space. 
Following previous work \cite{milioto2019rangenet++,wu2017squeezeseg,wu2018squeezesegv2}, we project all points in a scan to a $64 \times 2048$ image. If multiple points are projected to the same pixel on the 2D image, we keep the point with the largest distance. Following RangeNet21 and RangeNet53 in~\cite{milioto2019rangenet++}, we propose SqueezeSegV3-21 (SSGV3-21) and SqueezeSegV3-53 (SSGV3-53). The model architecture of SSGV3-21 and SSGV3-53 are similar to RangeNet21 and RangNet53~\cite{milioto2019rangenet++}, except that we replace regular convolution blocks with SAC blocks. Both models contain 5 stages, each of them has a different input resolution. In SSGV3-21, the 5 stages respectively contain 1, 1, 2, 2, 1 blocks and in SSGV3-53, the 5 stages contain 1, 2, 8, 8, 4 blocks, which are also same as  RangeNet21 and RangeNet53, respectively. 

We use the SGD optimizer to end-to-end train the whole model. During training, SSGV3-21 is trained with an initial learning rate of $0.01$, SSGV3-53 is trained with an initial learning rate of $0.005$. We use the warming up strategy to change the learning rate for 1 epoch. During inference, the original points will be projected and fed into SqueezeSegV3 to get a 2D prediction. Then we adopt the restoration operation to obtain the 3D prediction, as previous work~\cite{milioto2019rangenet++,wu2017squeezeseg,wu2018squeezesegv2}. 

\subsection{Comparing with Prior Methods}

\begin{table*}[!t]
\normalfont

\resizebox{1\textwidth}{!}{
\begin{tabular}{c c c c c c c c c c c c c c c c c c c c | c c}
 \hline
 \rotatebox{0}{Method} & \rotatebox{90}{car}&\rotatebox{90}{bicycle}&  \rotatebox{90}{motorcycle} &\rotatebox{90}{truck} &\rotatebox{90}{other-vehicle} &\rotatebox{90}{person}&\rotatebox{90}{bicyclist}&
\rotatebox{90}{Motorcyclist }&\rotatebox{90}{road}&\rotatebox{90}{parking}&\rotatebox{90}{sidewalk}&\rotatebox{90}{other-ground} &\rotatebox{90}{building}& \rotatebox{90}{fence} &\rotatebox{90}{vegetation}& \rotatebox{90}{trunk}&\rotatebox{90}{terrain}&\rotatebox{90}{pole}&\rotatebox{90}{traffic-sign} &\rotatebox{90}{mean IoU}& \rotatebox{90}{Scans/sec} \\
 \hline
 
 \rotatebox{0}{PNet~\cite{qi2017pointnet}} & 46.3 &1.3 &0.3& 0.1& 0.8& 0.2& 0.2& 0.0& 61.6& 15.8& 35.7& 1.4& 41.4& 12.9& 31.0& 4.6 &17.6& 2.4& 3.7& 14.6 & 2 \\

\rotatebox{0}{PNet++~\cite{qi2017pointnet++}} & 53.7 & 1.9 & 0.2 & 0.9 & 0.2 & 0.9 & 1.0 & 0.0 & 72.0 & 18.7 & 41.8 & 5.6 & 62.3 & 16.9 & 46.5 & 13.8 & 30.0 & 6.0 & 8.9 & 20.1 & 0.1 \\

\rotatebox{0}{SPGraph~\cite{landrieu2018large}} & 68.3 & 0.9 & 4.5 & 0.9 & 0.8 & 1.0 & 6.0 & 0.0 & 49.5 & 1.7 & 24.2 & 0.3 & 68.2 & 22.5 & 59.2 & 27.2 & 17.0 & 18.3 & 10.5 & 20.0 & 0.2 \\

\rotatebox{0}{SPLAT~\cite{su2018splatnet}} & 66.6 & 0.0 & 0.0 & 0.0 & 0.0 & 0.0 & 0.0 & 0.0 & 70.4 & 0.8 & 41.5 & 0.0 & 68.7 & 27.8 & 72.3 & 35.9 & 35.8 & 13.8 & 0.0 & 22.8 & 1 \\

\rotatebox{0}{TgConv~\cite{tatarchenko2018tangent}} & 86.8 & 1.3 & 12.7 & 11.6 & 10.2 & 17.1 & 20.2 & 0.5 & 82.9 & 15.2 & 61.7 & 9.0 & 82.8 & 44.2 & 75.5 & 42.5 & 55.5 & 30.2 & 22.2 & 35.9 & 0.3\\

\rotatebox{0}{RLNet~\cite{hu2019randla}} & 94.0 & 19.8 & 21.4 & 42.7 & 38.7 & 47.5 & 48.8 & 4.6 & 90.4 & 56.9 & 67.9 & 15.5 & 81.1 & 49.7 & 78.3 & 60.3 & 59.0 & 44.2 & 38.1 & 50.3 & 22\\
\hline
\rotatebox{0}{SSG~\cite{wu2017squeezeseg}} & 68.8 & 16.0 & 4.1 & 3.3 & 3.6 & 12.9 & 13.1 & 0.9 & 85.4 & 26.9 & 54.3 & 4.5 & 57.4 & 29.0 & 60.0 & 24.3 & 53.7 & 17.5 & 24.5 & 29.5 & \textbf{65}\\

\rotatebox{0}{SSG$\ddagger$~\cite{wu2017squeezeseg}} & 68.3 & 18.1 & 5.1 & 4.1 & 4.8 & 16.5 & 17.3 & 1.2 & 84.9 & 28.4 & 54.7 & 4.6 & 61.5 & 29.2 & 59.6 & 25.5 & 54.7 & 11.2 & 36.3 & 30.8 &53 \\

\rotatebox{0}{SSGV2~\cite{wu2018squeezesegv2}} & 81.8 & 18.5 & 17.9 & 13.4 & 14.0 & 20.1 & 25.1 & 3.9 & 88.6 & 45.8 & 67.6 & 17.7 & 73.7 & 41.1 & 71.8 & 35.8 & 60.2 & 20.2 & 36.3 & 39.7 & 50\\

\rotatebox{0}{SSGV2$\ddagger$~\cite{wu2018squeezesegv2}} & 82.7 & 21.0 & 22.6 & 14.5 & 15.9 & 20.2 & 24.3 & 2.9 & 88.5 & 42.4 & 65.5 & 18.7 & 73.8 & 41.0 & 68.5 & 36.9 & 58.9 & 12.9 & 41.0 & 39.6 & 39\\

\rotatebox{0}{RGN21~\cite{milioto2019rangenet++}} & 85.4 & 26.2 & 26.5 & 18.6 & 15.6 & 31.8 & 33.6 & 4.0 & 91.4 & 57.0 & 74.0 & 26.4 & 81.9 & 52.3 & 77.6 & 48.4 & 63.6 & 36.0 & 50.0 & 47.4 & 20 \\

\rotatebox{0}{RGN53~\cite{milioto2019rangenet++}} & 86.4 & 24.5 & 32.7 & 25.5 & 22.6 & 36.2 & 33.6 & 4.7 & \textbf{91.8} & 64.8 & 74.6 & \textbf{27.9} & 84.1 & 55.0 & 78.3 & 50.1 & 64.0 & 38.9 & 52.2 & 49.9 & 12\\


\rotatebox{0}{RGN53*~\cite{milioto2019rangenet++}}&91.4 &25.7& 34.4 &25.7& 23.0& 38.3& 38.8& 4.8& \textbf{91.8}& \textbf{65.0}& \textbf{75.2}& 27.8& 87.4& 58.6& 80.5& 55.1& 64.6& 47.9& 55.9& 52.2 & 11\\
\hline

\rotatebox{0}{SSGV3-21}&84.6&31.5&32.4&11.3&20.9&39.4&36.1&\textbf{21.3}&90.8&54.1&72.9&23.9&81.1&50.3&77.6&47.7&63.9&36.1&51.7&48.8 &16 \\
\rotatebox{0}{SSGV3-53} &  87.4&35.2&33.7&29.0&31.9&41.8&39.1&20.1&\textbf{91.8}&63.5&74.4&27.2&85.3&55.8&79.4&52.1&64.7&38.6&53.4&52.9 & 7 \\
\rotatebox{0}{SSGV3-21*}& 89.4&33.7&34.9&11.3&21.5&42.6&44.9&21.2&90.8&54.1&73.3&23.2&84.8&53.6&80.2&53.3&64.5&46.4&57.6&51.6& 15 \\
\rotatebox{0}{SSGV3-53*} & 92.5 &\textbf{38.7} & \textbf{36.5} & 29.6 & 33.0& 45.6 & 46.2 & 20.1&91.7 & 63.4 & 74.8 & 26.4 & \textbf{89.0} & \textbf{59.4} &\textbf{82.0}  & 58.7 & 65.4 &\textbf{49.6}  & \textbf{58.9} & \textbf{55.9}&6 \\

\hline
\rotatebox{0}{}
\end{tabular}}

\caption{IoU [\%] on test set (sequences 11 to 21). SSGV3-21 and SSGV3-53 are the proposed method. Their complexity corresponds to RangeNet21 and RangeNet53 respectively. 
* means KNN post-processing from RangeNet++~\cite{milioto2019rangenet++}, and $\ddagger$ means the CRF post-processing from SqueezeSegV2 used~\cite{wu2018squeezesegv2}. The first group reports PointNet-based methods. The second reports projection-based methods. The third include our results}
\label{tab:main-comparison}
\end{table*}

We compare two proposed models, SSGV3-21 and SSGV3-53, with previous published work~\cite{qi2017pointnet,qi2017pointnet++,landrieu2018large,su2018splatnet,tatarchenko2018tangent,wu2017squeezeseg,wu2018squeezesegv2,milioto2019rangenet++}. From Table \ref{tab:main-comparison}, we can see that the proposed SqueezeSegV3 models outperforms all the baselines. Compared with the previous state-of-the-art RangeNet53 \cite{milioto2019rangenet++}, SSGV3-53 improves the accuracy by 3.0 mIoU. Moreover, when we apply post-processing KNN refinement following~\cite{milioto2019rangenet++} (indicated as *), the proposed SSGV3-53* outperforms RangeNet53* by 3.7 mIoU and achieves the best accuracy in 14 out of 19 categories. Meanwhile, the proposed SSGV3-21 also surpasses RangeNet21 by 1.4 mIoU and the performance is close to RangeNet53* with post-processing. The advantages are more significant for smaller objects, as SSGV3-53* significantly outperforms RangeNet53* by 13.0 IoU, 10.0 IoU, 7.4 IoU and 15.3 IoU in categories of bicycle, other-vehicle, bicyclist and Motorcyclist respectively. 

In terms of speed, SSGV3-21 (16 FPS) is closet RangeNet21 (20 FPS). Even though SSGV3-53 (7 FPS) is slower than RangeNet53 (12 FPS), note that our implementation of SAC is primitive and it can be optimized to achieve further speedup.
In comparison, PointNet-based methods~\cite{qi2017pointnet,qi2017pointnet++,landrieu2018large,su2018splatnet,tatarchenko2018tangent} do not perform well in either accuracy and speed except RandLA-Net~\cite{hu2019randla} which is a new efficient and effective work.


\subsection{Ablation Study}
\label{sec:ablation}
We conduct ablation studies to analyze the performance of SAC with different configurations. Also, we compare it with other related operators to show its effectiveness. To facilitate fast training and experiments, we shrink the LiDAR images to $64 \times 512$, and use the shallower model of SSGV3-21 as the starting point. 
We evaluate the accuracy directly on the projected 2D image, instead of the original 3D points, to make the evaluation faster. We train the models in this section on the training set of SemanticKITTI and report the accuracy on the validation set. We study different variations of SAC, input kernel sizes, and other techniques used in SqueezeSegV3.


\begin{table}[!t]
\resizebox{1\textwidth}{!}{
\footnotesize
\centering
\small
\begin{tabular}{ c c c c c c c c c c}
 \hline
 Method &  Baseline & SAC-S & SAC-IS  & SAC-SK &SAC-ISK& PAC~\cite{su2019pixel}  & SE~\cite{hu2018squeeze} & CBAM~\cite{woo2018cbam}&CAM~\cite{wu2018squeezesegv2}  \\
 \hline
 \hline
 mIoU& 44.0 & 44.9 & 44.0 & 45.4 &46.3& 45.2& 44.2& 44.8 & 42.1 \\
 Accuracy& 86.8  & 87.6 & 86.9  & 88.2 & 88.6&88.2& 87.0& 87.5&85.8\\
\hline 
\end{tabular}
}

\caption{mIoU [\%] and Accuracy [\%] for variants of spatially-adaptive convolution}
\label{tab:sac-variants}
\end{table}

\noindent\textbf{Variants of spatially-adaptive convolution:} As shown in Figure~\ref{fig:form} \& \ref{fig:others}, spatially-adaptive convolution can have many variation. Some variants are equivalent to or similar with methods proposed by previous papers, including squeeze-and-excitation (SE) \cite{hu2018squeeze}, convolutional block attention maps (CBAM) \cite{woo2018cbam}, pixel-adaptive convolution (PAC) \cite{su2019pixel}, and context-aggregation module (CAM) \cite{wu2018squeezesegv2}. To understand the effectiveness of SAC variants and previous methods, we swap them into SqueezeSegV3-21. The results are reported in Table.~\ref{tab:sac-variants}. 

It can be seen that SAC-ISK significantly outperforms all the other settings in term of mIoU. CAM and SAC-IS have the worst performance, which demonstrates the importance of the attention on the kernel dimension. Squeeze-and-excitation (SE) also does not perform well, since SE is not spatially-adaptive, and the global average pooling used in SE ignores the feature distribution shift across the LiDAR image. In comparison, CBAM~\cite{woo2018cbam} improves the baseline by 0.8 mIoU. Unlike SE, it also adapts the input feature spatially. This comparison shows that being spatially-adaptive is crucial for processing LiDAR images. Pixel-adaptive convolution (PAC) is similar to the SAC variant of SAC-SK, except that PAC uses a Gaussian function to compute the kernel-wise attention. Experiments show that the proposed SAC-SK slightly outperforms SAC-SK, possibly because SAC-SK adopts a more general and learnable convolution to compute the attention map. Comparing SAC-S and SAC-IS, adding the input channel dimension does not improve the performance.

\begin{table}[!t]
\footnotesize

\centering
\begin{tabular}{ c c c c c c c }
 \hline
Kernel size & baseline & $1 \times 1$ & $3 \times 3$ & $5 \times 5$ & $7 \times 7$ \\
 \hline
 \hline
 mIoU&44.0&45.5 &44.5 &45.4 & 46.3\\
 Accuracy& 86.8& 88.4& 87.6&88.2 &88.6\\
\hline
\end{tabular}

\caption{mIoU [\%] and Accuracy [\%] for different convolution kernel sizes for coordinate map}
\label{tab:kernel}
\end{table}
\noindent\textbf{Kernel Sizes of SAC:}  We use a one-layer convolution to compute the attention map for SAC. However, what should be the kernel size for this convolution? A larger kernel size makes sure that it can capture spatial information around, but it also costs more parameters and MACs. To examine the influence of kernel size, we use different kernel sizes in the SAC convolution. As we can see in Table~\ref{tab:kernel}, a 1x1 convolution provides a very strong result that is better than its 3x3 and 5x5 counterparts. 7x7 convolution performs the best.  

\noindent\textbf{The effectiveness of other techniques:} In addition to SAC, we also introduce several new techniques to SqueezeSegV3, including removing the last two downsample layers and multi-layer loss. We start from the baseline of RangeNet21. 
First, we remove downsampling layers and reduce the channel sizes of the last two stages to 256 to keep the MACS the same. The performance improves by 3.9 mIoU. After adding the multi-layer loss, the mIoU increases by another 1.5\%. Based on the above techniques, adding SAC-ISK further boost mIoU by 2.3\%.

\begin{table}[!t]

\footnotesize
\centering
\begin{tabular}{c c c c c c }
 \hline
method& Baseline & +DS removal & +Multi-layer loss & +SAC-ISK & \\
 \hline
 \hline
 mIoU & 38.6&42.5 (+3.9) & 44.0 (+1.5)& 46.3 (+2.3)\\
 Accuracy & 84.7 & 86.2 (+1.5) &86.8 (+1.4) &88.6 (+1.8)\\
\hline
\end{tabular}

\caption{mIoU [\%] and Accuracy [\%] with downsampling removal, multi-layer loss, and spatially-adaptive convolution}
\label{tab:other}
\end{table}



\clearpage
%
%
\bibliographystyle{splncs04}
\bibliography{egbib}

\clearpage
\section{Appendix}

In this appendix, we provide the pseudo code implementation for SAC variants discussed in Section~\ref{sec:sac-variant}, including SAC-S, SAC-IS, SAC-SK and SAC-ISK. 

\begin{figure}[t!]
     \centering
     \begin{subfigure}[b]{0.75\textwidth}
         \centering
         \includegraphics[width=\textwidth]{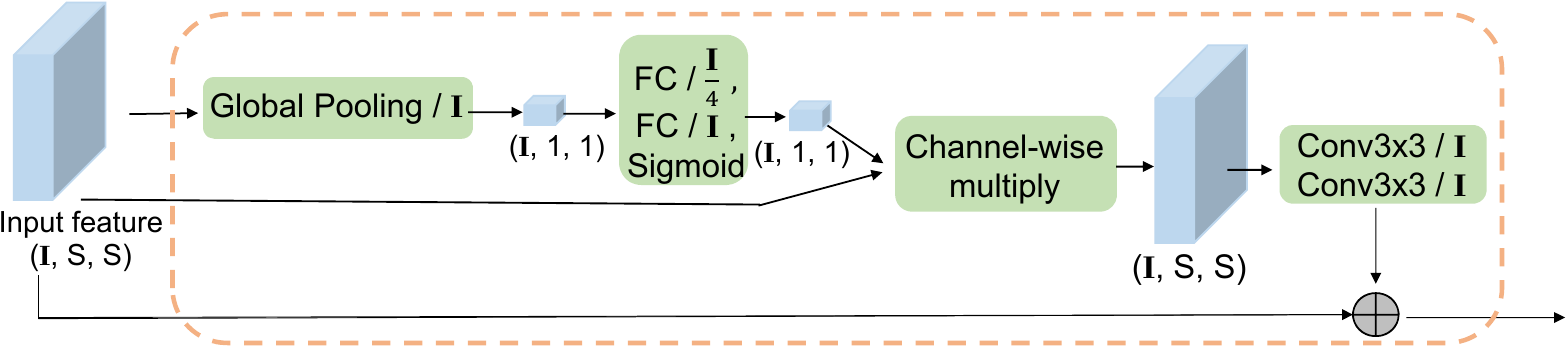}
         \caption{ Squeeze-and-Excitation~\cite{hu2018squeeze}}
         \hrule
         \label{fig:SE}
     \end{subfigure}
     \hfill
          \begin{subfigure}[b]{0.75\textwidth}
         \centering
         \includegraphics[width=\textwidth]{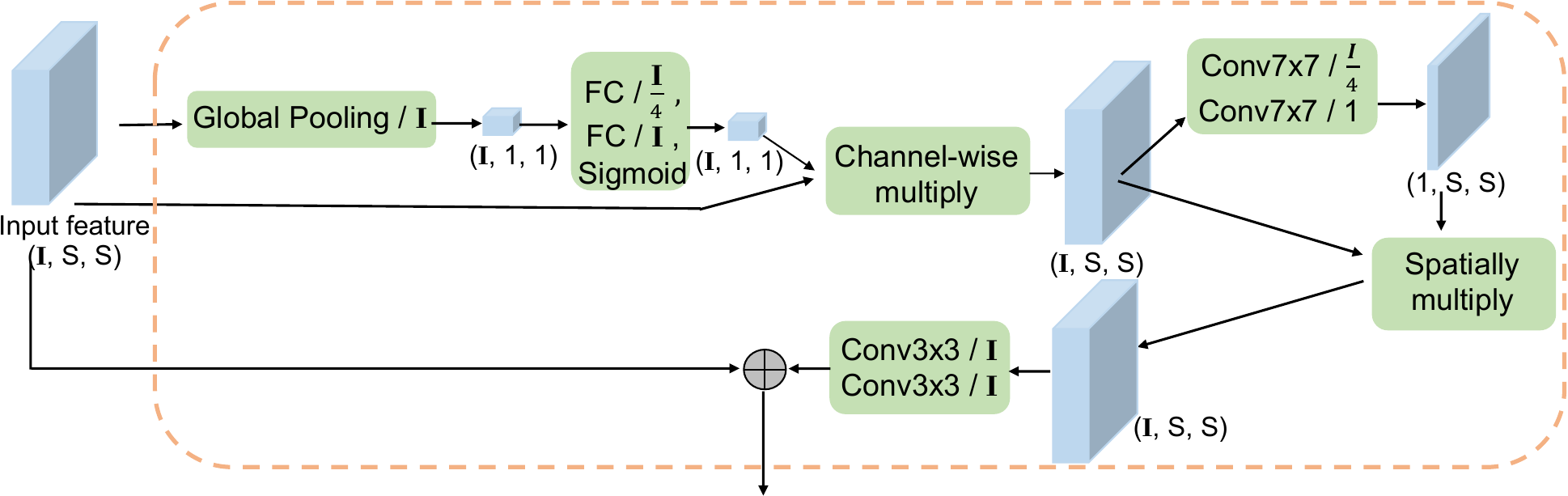}
         \caption{CBAM~\cite{woo2018cbam}.}
         \hrule
         \label{fig:CBAM}
     \end{subfigure}
          \hfill
          \begin{subfigure}[b]{0.75\textwidth}
         \centering
         \includegraphics[width=\textwidth]{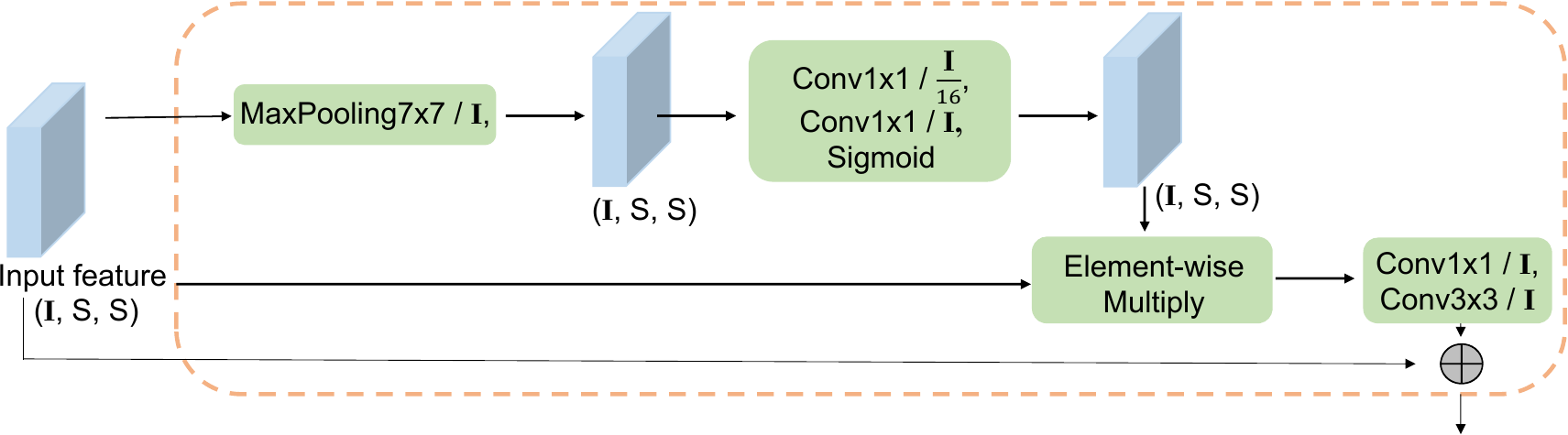}
         \caption{CAM~\cite{wu2018squeezesegv2}}
         \hrule
         \label{fig:CAM}
     \end{subfigure}
               \hfill
          \begin{subfigure}[b]{0.75\textwidth}
         \centering
         \includegraphics[width=\textwidth]{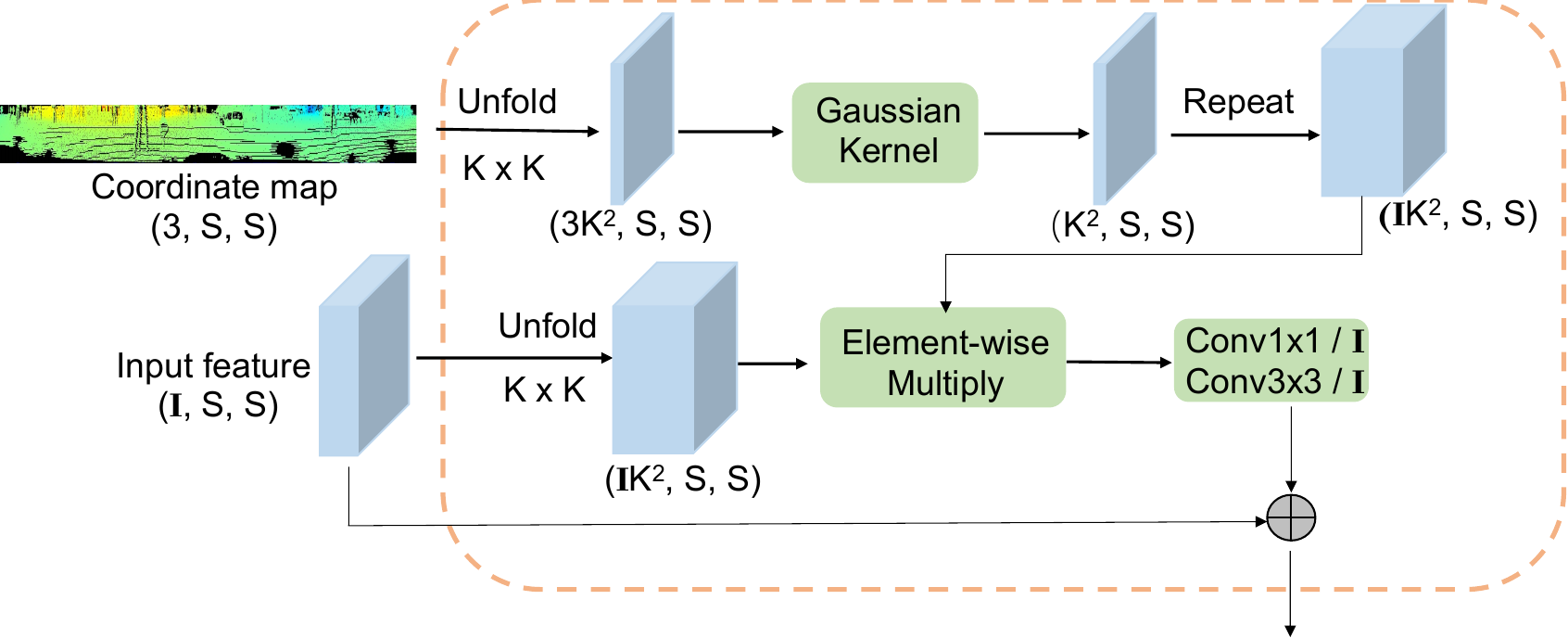}
         \caption{PAC~\cite{su2019pixel}}
         \label{fig:PAC}
     \end{subfigure}
     \hfill
     
    \caption{Variants of spatially-adaptive convolution from previous work.}
    \label{fig:others}
\end{figure}

\begin{minted}[mathescape]{python}
""" The input of each function is input_feature and coordinate map.
input_feature (N, C, H, W), coordinate_map (N, 3, H, W)
output_feature (N, C, H, W) """
\end{minted}
\begin{minted}[mathescape]{python}

def SAC_S(input_feature, coordi_map):
# Note: Pseudo code for SAC-S.
  attention_map = Conv_attention7x7(coordin_map) # (N, 1, H, W)
  input_feature = input_feature * attention_map # (N, C, H, W)
  feature = Conv_feature3x3(input_feature) # (N, C, H, W)
  output_feature = feature+input_feature # (N, C, H, W)
  return output_feature # (N, C, H, W)
\end{minted}

\begin{minted}[mathescape]{python}
def SAC_IS(input_feature, coordi_map):
# Note: Pseudo code for SAC-IS.
  attention_map = Conv_attention7x7(coordin_map) # (N, C, H, W)
  input_feature = input_feature * attention_map # (N, C, H, W)
  feature = Conv_feature3x3(input_feature) # (N, C, H, W)
  output_feature = feature+input_feature # (N, C, H, W)
  return output_feature # (N, C, H, W)
\end{minted}

\begin{minted}[mathescape]{python}
def SAC_SK(input_feature, coordi_map):
# Note: Pseudo code for SAC-SK.
  unfold_feature = unfold(input_feature, kernel_size=K, 
  padding=K//2) # (N, C*K*K, H, W)
  attention_map = Conv_attention7x7(coordin_map) # (N, K*K, H, W)
  attention_map = attention_map.repeat(1, C, 1, 1) # (N, C*K*K, H, W)
  input_feature = input_feature * attention_map # (N, C*K*K, H, W)
  feature = Conv_feature1x1(input_feature) # (N, C, H, W)
  feature = Conv_feature3x3(feature) # (N, C, H, W)
  output_feature = feature+input_feature # (N, C, H, W)
  return output_feature # (N, C, H, W)
    
\end{minted}

\begin{minted}[mathescape]{python}
def SAC_ISK(input_feature, coordi_map):
# Note: Pseudo code for SAC-ISK.
  unfold_feature = unfold(input_feature, kernel_size=K, 
  padding=K//2) # (N, C*K*K, H, W )
  attention_map = Conv_attention7x7(coordin_map) # (N, C*K*K, H, W)
  input_feature = input_feature * attention_map # (N, C*K*K, H, W)
  feature = Conv_feature1x1(input_feature) # (N, C, H, W)
  feature = Conv_feature3x3(feature) # (N, C, H, W)
  output_feature = feature+input_feature # (N, C, H, W)
  return output_feature # (N, C, H, W)
\end{minted}

\end{document}